%% file: main.tex
\documentclass[10pt,twocolumn,letterpaper]{article}
\usepackage{multirow}
\usepackage{makecell}
\usepackage[ruled,vlined]{algorithm2e}
\usepackage[pagenumbers]{iccv} 


\definecolor{iccvblue}{rgb}{0.21,0.49,0.74}
\definecolor{urlcolor}{RGB}{255,105,180}
\usepackage[pagebackref,breaklinks,colorlinks,allcolors=iccvblue,urlcolor=urlcolor]{hyperref}

\def\paperID{4793} 
\def\confName{ICCV}
\def\confYear{2025}

\title{UniCombine: Unified Multi-Conditional Combination \\ with Diffusion Transformer}

\author{
\vspace{-2em}\\
Haoxuan Wang$^{1 \dagger}$, Jinlong Peng$^{2 \dagger}$, Qingdong He$^{2}$, Hao Yang$^{3}$, Ying Jin$^{1}$, Jiafu Wu$^{2}$,\\
Xiaobin Hu$^{2}$, Yanjie Pan$^{1}$, Zhenye Gan$^{2}$, Mingmin Chi$^{1*}$, Bo Peng$^{4*}$, Yabiao Wang$^{2,5*}$ \\
\small
{\textit{$^{1}$Fudan University,} 
\textit{$^{2}$Tencent Youtu Lab,} 
\textit{$^{3}$Shanghai Jiao Tong University,}
\textit{$^{4}$Shanghai Ocean University}
\textit{$^{5}$Zhejiang University}
}\\
\\
\url{https://github.com/Xuan-World/UniCombine}
}

\begin{document}

\twocolumn[{ 
\maketitle 
\begin{center}
\centering 
\vspace{-20px}
\captionsetup{type=figure}
\includegraphics[width=\textwidth]{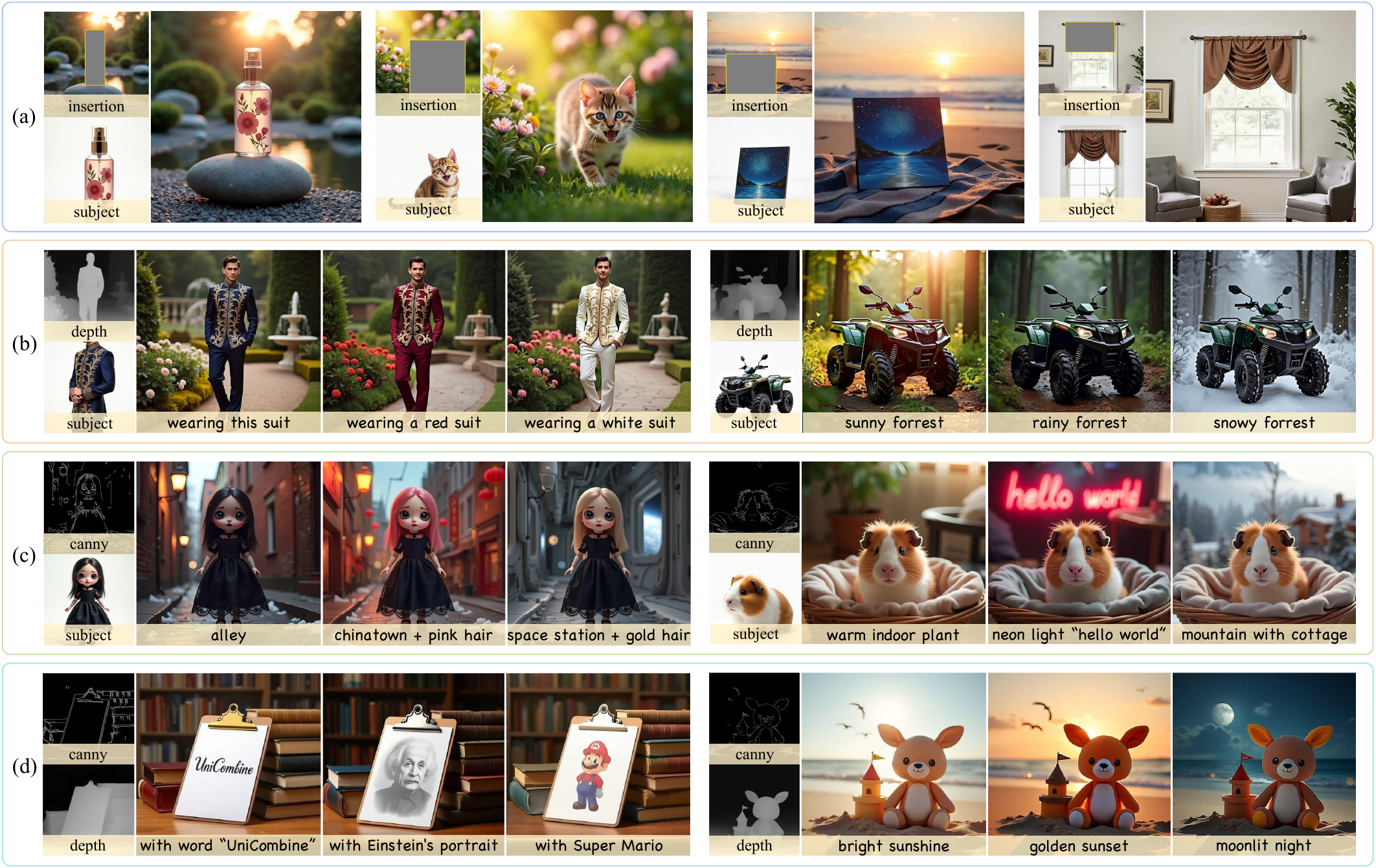}  
\vspace{-1.5em}
\caption{Fantastic results of our proposed UniCombine on multi-conditional controllable generation: (a) Subject-Insertion task. (b) and (c) Subject-Spatial task. (d) Multi-Spatial task. Our unified framework effectively handles any combination of input conditions and achieves remarkable alignment with all of them, including but not limited to text prompts, spatial maps, and subject images.
}
\label{fig:first_image}
\end{center} 
}]
\renewcommand{\thefootnote}{}
\footnotetext[1]{$\dagger~$Equal contribution.}
\footnotetext[2]{$*~$Corresponding author.}
\input{sec/0_abstract}    
\input{sec/1_intro}
\input{sec/2_relate}
\input{sec/3_method}
\input{sec/4_experiment}

\input{sec/5_conclusion}

{
    \small
    \bibliographystyle{ieeenat_fullname}
    \bibliography{main}
}

\input{sec/6_supplement}

\end{document}

%% file: sec/0_abstract.tex
\begin{abstract}
With the rapid development of diffusion models in image generation, the demand for more powerful and flexible controllable frameworks is increasing. Although existing methods can guide generation beyond text prompts, the challenge of effectively combining multiple conditional inputs while maintaining consistency with all of them remains unsolved. To address this, we introduce \textbf{UniCombine}, a DiT-based multi-conditional controllable generative framework capable of handling any combination of conditions, including but not limited to text prompts, spatial maps, and subject images. Specifically, we introduce a novel Conditional MMDiT Attention mechanism and incorporate a trainable LoRA module to build both the training-free and training-based versions. Additionally, we propose a new pipeline to construct SubjectSpatial200K, the first dataset designed for multi-conditional generative tasks covering both the subject-driven and spatially-aligned conditions. Extensive experimental results on multi-conditional generation demonstrate the outstanding universality and powerful capability of our approach with state-of-the-art performance.

\end{abstract}

%% file: sec/1_intro.tex
\section{Introduction}
\label{sec:intro}

With the advancement of diffusion-based \cite{ddpm,ddim} text-to-image generative technology, a series of single-conditional controllable generative frameworks like ControlNet \cite{controlnet}, T2I-Adapter \cite{T2Iadapter}, IP-Adapter \cite{IPadapter}, and InstantID \cite{instantid} have expanded the scope of the control signals from text prompts to image conditions. It allows users to control more plentiful aspects of the generated images, such as layout, style, characteristics, etc. 
These conventional approaches are specifically designed for the UNet \cite{unet} backbone of Latent Diffusion Models (LDM) \cite{ldm} with dedicated control networks. 
Besides, some recent approaches, such as OminiControl \cite{ominicontrol}, integrate control signals into the Diffusion Transformer (DiT) \cite{sd3,flux} architecture, which demonstrates superior performance compared to the UNet in LDM.

Although the methods mentioned above have achieved a promising single-conditional performance, the challenge of multi-conditional controllable generation is still unsolved. Previous multi-conditional generative methods like UniControl \cite{unicontrol} and UniControlNet \cite{unicontrolnet} are generally restricted to handling spatial conditions like Canny or Depth maps and fail to accommodate subject conditions, resulting in limited applicable scenarios. Despite the recently proposed Ctrl-X \cite{ctrlx} features controlling structure and appearance together, its performance is unsatisfactory and supports only a limited combination of conditions.

Moreover, we assume that many existing generative tasks can be viewed as a multi-conditional generation, such as virtual try-on \cite{catvton, jiang2024fitdit}, object insertion \cite{objectmate, chen2024mureobjectstitch}, style transfer \cite{csgo, hu2023stroke, peng2024frih}, spatially-aligned customization \cite{ctrlx, instantfamily, kong2024anymaker, li2024tuning}, etc. Consequently, there is a need for a unified framework to encompass these generative tasks in a way of multi-conditional generation. This framework should ensure consistency with all input constraints, including subject ID preservation, spatial structural alignment, background coherence, and style uniformity.

To achieve this, we propose UniCombine, a powerful and universal framework that offers several key advantages: 
\textbf{Firstly}, our framework is capable of simultaneously handling any combination of conditions, including but not limited to text prompts, spatial maps, and subject images. 
Specifically, we introduce a novel Conditional MMDiT Attention mechanism and incorporate a trainable Denoising-LoRA module to build both the training-free and training-based versions. By integrating multiple pre-trained Condition-LoRA module weights into the conditional branches, UniCombine achieves excellent training-free performance, which can be improved further after training on the task-specific multi-conditional dataset.
\textbf{Secondly}, due to the lack of a publicly available dataset for multi-conditional generative tasks, we build the SubjectSpatial200K dataset to serve as the training dataset and the testing benchmark. Specifically, we generate the subject grounding annotations and spatial map annotations for all the data samples from Subjects200K \cite{ominicontrol} and therefore formulate our SubjectSpatial200K dataset.
\textbf{Thirdly}, our UniCombine can achieve many unprecedented multi-conditional combinations, as shown in \cref{fig:first_image}, such as combining a reference subject image with the inpainting area of a background image or with the layout guidance of a depth (or canny) map while imposing precise control via text prompt. 
Furthermore, extensive experiments on Subject-Insertion, Subject-Spatial, and Multi-Spatial conditional generation demonstrate the outstanding universality and powerful capability of our method against other existing specialized approaches. 

In summary, we highlight our contributions as follows:
\begin{itemize}[left=0pt]
    \item We present UniCombine, a DiT-based multi-conditional controllable generative framework capable of handling any combination of conditions, including but not limited to text prompts, spatial maps, and subject images.
    
    \item  We construct the SubjectSpatial200K dataset, which encompasses both subject-driven and spatially-aligned conditions for all text-image sample pairs. It addresses the absence of a publicly available dataset for training and testing multi-conditional controllable generative models.
    
    \item We conduct extensive experiments on Subject-Insertion, Subject-Spatial, and Multi-Spatial conditional generative tasks. The experimental results demonstrate the state-of-the-art performance of our UniCombine, which effectively aligns with all conditions harmoniously.
\end{itemize}

\begin{figure*}[t]
    \centering
    \includegraphics[width=\textwidth]{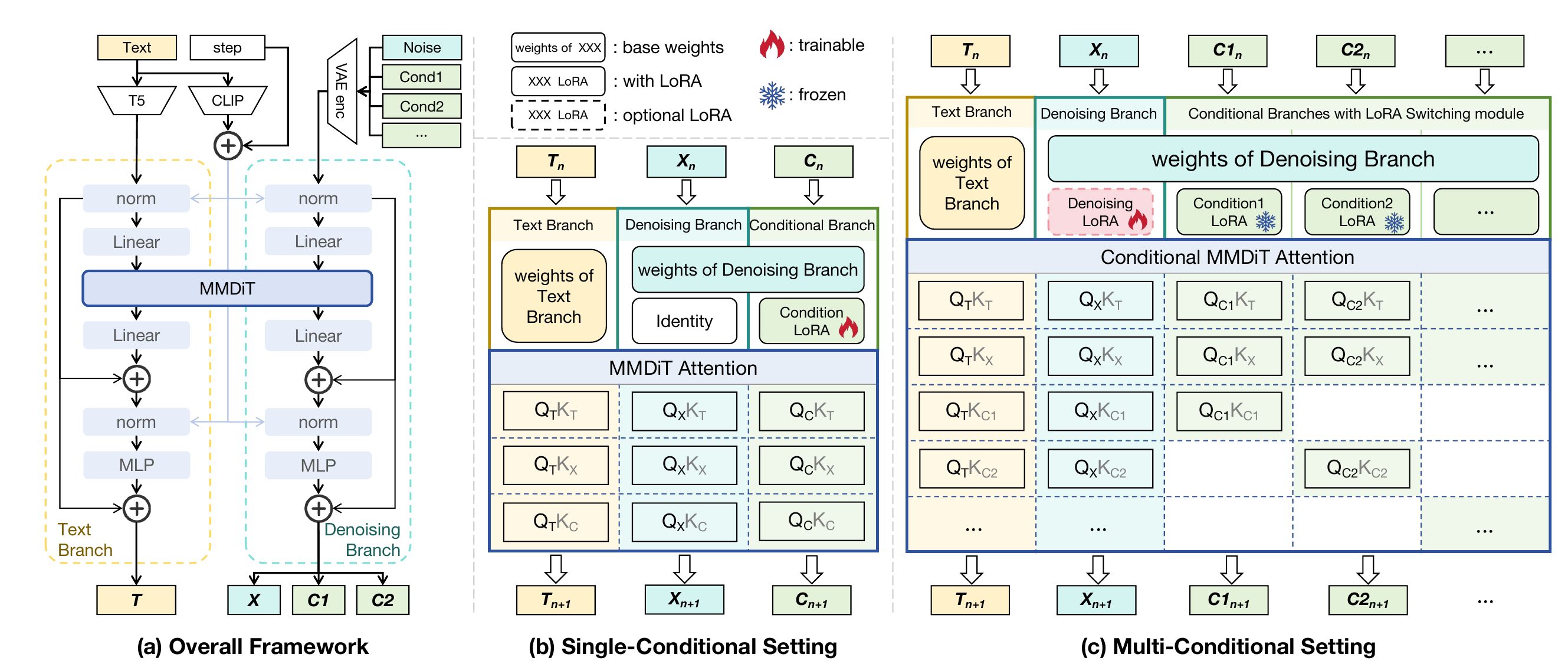}
    \vspace{-1.5em}
    \caption{Overview of our proposed UniCombine. (a) The overall framework. We regard the MMDiT-based diffusion models as consisting of the text branch and the denoising branch. Based on it, our UniCombine introduces multiple conditional branches to process the input conditions. (b) The single-conditional setting of our UniCombine. It is equivalent to OminiControl \cite{ominicontrol} which is a special case of our proposed UniCombine framework under a single-conditional setting. (c) The multi-conditional setting of our UniCombine. Our LoRA Switching module adaptively activates the pre-trained Condition-LoRA modules on the weights of the denoising branch according to the conditional types. The proposed Conditional MMDiT Attention mechanism is used to replace the original MMDiT Attention mechanism for handling the unified multi-conditional input sequence. Whether to load the optional Denoising-LoRA module is the difference between the training-free and training-based versions.}
    \label{fig:method}
    \vspace{-1em}
\end{figure*}

%% file: sec/2_relate.tex
\section{Related Work}
\label{sec:rel}

\subsection{Diffusion-Based Models}
Diffusion-based \cite{ddim,ddpm} models have demonstrated superior performance than GAN-based \cite{gan} ones across various domains, including controllable generation  \cite{controlnet, T2Iadapter, IPadapter, instantid, jin2024dualanodiff}, image editing  \cite{prompt2promt, nulltextInversion, RFinversion}, customized generation  \cite{textualInverion, dreambooth, customDiffusion}, object insertion  \cite{pbe, objectStitch, anydoor}, mask-guided inpainting  \cite{imageneditor, powerpaint, brushnet}, and so on. These breakthroughs begin with the LDM  \cite{ldm} and are further advanced with the DiT  \cite{dit} architecture. The latest text-to-image generative models, SD3  \cite{sd3} and FLUX  \cite{flux}, have attained state-of-the-art results by employing the Rectified Flow \cite{fm, rf} training strategy, the RPE \cite{rotaryPE} positional embedding and the Multi-Modal Diffusion Transformer (MMDiT) \cite{sd3} architecture.

\subsection{Controllable Generation}
Controllable generation allows for customizing the desired spatial layout, filter style, or subject appearance in the generated images.
A series of methods such as ControlNet \cite{controlnet}, T2I-Adapter \cite{T2Iadapter}, GLIGEN \cite{gligen}, and ZestGuide \cite{ZestGuide} successfully introduce the spatial conditions into controllable generation, enabling models to control the spatial layout of generated images. Another series of methods, such as IP-Adapter \cite{IPadapter}, InstantID \cite{instantid}, BLIP-Diffusion \cite{blipdiffusion}, and StyleDrop \cite{styledrop} incorporate the subject conditions into controllable generation, ensuring consistency between generated images and reference images in style, characteristics, subject appearance, etc. 
To unify these two tasks, OminiControl \cite{ominicontrol} proposes a novel MMDiT-based controllable framework to handle various conditions with a unified pipeline. Unfortunately, it lacks the capability to control generation with multiple conditions. To this end, we propose UniCombine, which successfully extends this framework to multi-conditional scenarios.

\subsection{Multi-Conditional Controllable Generation}
As controllable generation advances, merely providing a single condition to guide the image generation no longer satisfies the needs. As a result, research on multi-conditional controllable generation has emerged. Existing methods like UniControl \cite{unicontrol}, UniControlNet \cite{unicontrolnet} and Cocktail \cite{cocktail} exhibit acceptable performance when simultaneously leveraging multiple spatial conditions for image generation. 
However, there is a lack of multi-conditional generative models that support utilizing both spatial conditions
and subject conditions
to guide the generative process together. Although the recently proposed method Ctrl-X \cite{ctrlx} features controlling the appearance and structure simultaneously, its performance remains unsatisfactory with a limited combination of conditions and it is not compatible with the Diffusion Transformer architecture.
To address the aforementioned limitations,  we propose UniCombine to enable the flexible combination of various control signals. 

%% file: sec/3_method.tex
\section{Method}
\label{sec:method}

\subsection{Preliminary}
In this work, we mainly explore the latest generative models that utilize the Rectified Flow (RF) \cite{fm, rf} training strategy and the MMDiT \cite{sd3} backbone architecture, like FLUX \cite{flux} and SD3 \cite{sd3}. 
For the source noise distribution $X_0\!\sim\!p_{\text{noise}}$ and the target image distribution $X_1\!\sim\!p_{\text{data}}$, the RF defines a linear interpolation between them as $X_t = (1-t)X_0 + tX_1$ for $t\in[0,1]$. The training objective is to learn a time-dependent vector field $v_t(X_t, t; \theta)$ that describes the trajectory of the ODE $dX_t = v_t(X_t, t; \theta)dt$. Specifically, $v_t(X_t, t; \theta)$ is optimized to approximate the constant velocity $X_1 - X_0$, leading to the loss function as \cref{eq:rf}.

{
\vspace{-1em}
\footnotesize
\begin{equation}
    \mathcal{L}_{\text{RF}}(\theta) = \mathbb{E}_{X_1 \sim p_{\text{data}}, X_0\sim p_{\text{noise}}, t\sim U[0,1]} \Bigl[ \| (X_1 \! -  \! X_0) - v_t(X_t, t; \theta) \|^2 \Bigr]
\label{eq:rf}
\end{equation}
}

In this paper, we propose a concept of \textit{branch} to differentiate the processing flows of input embeddings from different modalities in MMDiT-based models. As shown in \cref{fig:method} (a), instead of the single-branch architecture \cite{ldm} where the text prompt is injected into the denoising branch via cross-attention, MMDiT uses two independent transformers to construct the text branch and the denoising branch.
Based on it, OminiControl \cite{ominicontrol} incorporates a Condition-LoRA module onto the weights of the denoising branch to process the input conditional embedding, thus forming its Conditional Branch, as depicted in \cref{fig:method} (b). \textit{It is worth noting that}, OminiControl \cite{ominicontrol} can be regarded as a special case of our proposed UniCombine framework under the single-conditional setting. It provides the pre-trained Condition-LoRA modules to meet the need for our multi-conditional settings. In the single-conditional setting, the text branch embedding \(T\), the denoising branch embedding \(X\), and the conditional branch embedding \(C\) are concatenated to form a unified sequence \([T; X; C]\) to be processed in the MMDiT Attention mechanism.

\subsection{UniCombine}
\label{sec:unicombine}
Building upon the MMDiT-based text-to-image generative model FLUX \cite{flux}, we propose UniCombine, a multi-conditional controllable generative framework consisting of various conditional branches. Each conditional branch is in charge of processing one conditional embedding, thus forming a unified embedding sequence $S$ as presented in \cref{eq:seq}. 

{
\vspace{-0.5em}
\footnotesize
\begin{equation}
    S =  [T; X; C_1; \dots; C_N]
\label{eq:seq}
\end{equation}
}

Given that the single-conditional setting of our UniCombine is equivalent to OminiControl \cite{ominicontrol}, we only focus on the multi-conditional setting in this section.
\textit{Firstly}, we introduce a LoRA Switching module to manage multiple conditional branches effectively.
\textit{Secondly}, we introduce a novel Conditional MMDiT Attention mechanism to process the unified sequence $S$ in the multi-conditional setting.
\textit{Thirdly}, we present an insight analysis of our training-free strategy,  which leverages the pre-trained Condition-LoRA module weights to perform a training-free multi-conditional controllable generation.
\textit{Lastly}, we present a feasible training-based strategy, which utilizes a trainable Denoising-LoRA module to enhance the performance further after training on a task-specific multi-conditional dataset.

\noindent \textbf{LoRA Switching Module.}
Before denoising with multiple input conditions, the Condition-LoRA modules pre-trained under single-conditional settings should be loaded onto the weights of the denoising branch, like $[ CondLoRA_{1}, CondLoRA_{2}, \dots]$. Then the LoRA Switching module determines which one of them should be activated according to the type of input conditions, forming a one-hot gating mechanism [0,1,0,…,0], as shown in \cref{fig:method} (c). 
Subsequently, different conditional branches with different activated Condition-LoRA modules are used for processing different conditional embeddings, resulting in a minimal number of additional parameters introduced for different conditions.
Unlike the single-conditional setting in \cref{fig:method} (b), which only needs loading LoRA modules, the LoRA Switching module in  \cref{fig:method} (c) enables adaptive selection among multiple LoRA modules to provide the matching conditional branches for each conditional embeddings, granting our framework greater flexibility and adaptability to handle diverse conditional combinations.

\noindent \textbf{Conditional MMDiT Attention.}
After concatenating the output embeddings from these \(N\) conditional branches, the unified sequence $S$ cannot be processed through the original MMDiT Attention mechanism due to two major challenges: 
(1) The computational complexity scales quadratically as \(O(N^2)\) with respect to the number of conditions, which becomes especially problematic when handling multiple high-resolution conditions.
(2) When performing MMDiT Attention on the unified sequence $S$, different condition signals interfere with each other during the attention calculation, making it difficult to effectively utilize the pre-trained Condition-LoRA module weights for the denoising process. 

To address these challenges, we introduce a novel Conditional MMDiT Attention mechanism (CMMDiT Attention) as depicted in \cref{fig:method} (c) to replace the original MMDiT Attention. Instead of feeding the entire unified sequence $S$ into the MMDiT Attention at once, CMMDiT Attention follows distinct computational mechanisms according to which branch is serving as queries. The core idea is that the branch serving as a query aggregates the information from different scopes of the unified sequence $S$ depending on its type. Specifically, when the denoising branch $X$ and the text branch $T$ serve as queries, their scope of keys and values correspond to the entire unified sequence $S$, granting them a global receptive field and the ability to aggregate information from all conditional branches. 
In contrast, when the conditional branches $C_i$ serve as queries, their receptive fields do not encompass one another. Their scope of keys and values are restricted to the subsequence $S_i$ as presented in \cref{eq:seq_i}, which prevents feature exchange and avoids information entanglement between different conditions. 

{
\vspace{-0.5em}
\footnotesize
\begin{equation}
    S_i = [T; X; C_{i}]
\label{eq:seq_i}
\end{equation}
}

Furthermore, the CMMDiT Attention reduces computational complexity from \(O(N^2)\) to \(O(N)\) as the number of conditions increases, making it more scalable.

\noindent \textbf{Training-free Strategy.}
The following analyses provide a detailed explanation of why our UniCombine is capable of seamlessly integrating and effectively reusing the pre-trained Condition-LoRA module weights to tackle multi-conditional challenges in a training-free manner. 

On the one hand, when the conditional embeddings $C_i$ serve as queries in CMMDiT, they follow the same attention computational paradigm as in the MMDiT of single-conditional settings, as indicated in \cref{eq:CasQ}.

{
\vspace{-1em}
\footnotesize
\begin{align}
    & \mathrm{CMMDiT} (Q=C_i^q,K=[T^k,X^k,C_i^k],V=[T^v,X^v,C_i^v]) \notag \\
    =& \mathrm{MMDiT} (Q=C^q,K=[T^k,X^k,C^k],V=[T^v,X^v,C^v])
\label{eq:CasQ}
\end{align}
}

\noindent This consistent computational paradigm enables the conditional branches to share the same feature extraction capability between the multi-conditional setting and the single-conditional setting.

On the other hand, when the denoising embedding $X$ and the text prompt embedding $T$ serve as queries in CMMDiT, their attention computational paradigm diverges from the single-conditional settings.
As illustrated in \cref{eq:NasQ}, when the denoising embedding $X$ is used as a query for attention computation with multiple conditional embeddings in CMMDiT, the attention score matrix is computed between $X$ and all the conditional embeddings.

{
\vspace{-1em}
\footnotesize
\begin{align}
    & \mathrm{CMMDiT}(Q=X^q,K/V=[X^{k/v},T^{k/v},C_1^{k/v},\ldots,C_N^{k/v}]) \notag \\
    =& \operatorname{softmax}(\frac{1}{\sqrt{dim}}X^q[X^k,T^k,C_1^k,\ldots,C_N^k]^\top)[X^v,T^v,C_1^v,\ldots,C_N^v]
\label{eq:NasQ}
\end{align}

}

\noindent It allows $X$ to extract and integrate information from each of the conditional embeddings separately and fusion them. This divide-and-conquer computational paradigm enables the text branch and denoising branch to fuse the conditional features effectively.


By leveraging the computational paradigms mentioned above, our UniCombine is able to perform a training-free multi-conditional controllable generation with the pre-trained Condition-LoRA modules.

\begin{figure}[t]
    \centering
    \includegraphics[width=\linewidth]{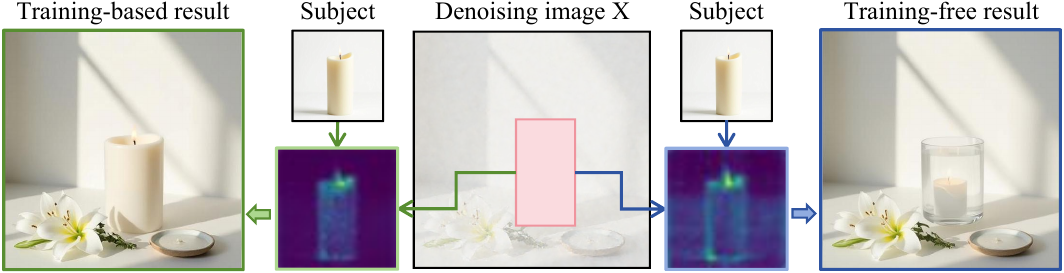}
    \vspace{-1.5em}
    \caption{Average X $\!\rightarrow\!$ Subject cross-attention map of the insertion area.}
    \label{fig:attnmap}
    \vspace{-1em}
\end{figure}

\noindent \textbf{Training-based Strategy.}
However, due to the lack of training, solely relying on the softmax operation in \cref{eq:NasQ} to balance the attention score distribution across multiple conditional embeddings may result in an undesirable feature fusion result, making our training-free version unsatisfactory in some cases. 
To address this issue, we introduce a trainable Denoising-LoRA module within the denoising branch to rectify the distribution of attention scores in \cref{eq:NasQ}. During training, we keep all the Condition-LoRA modules frozen to preserve the conditional extracting capability and train the Denoising-LoRA module solely on the task-specific multi-conditional dataset, as shown in \cref{fig:method} (c).  After training, the denoising embedding $X$ learns to better aggregate the appropriate information during the CMMDiT Attention operation. As presented in \cref{fig:attnmap}, the average X $\!\rightarrow\!$ Subject attention map within the inpainting area is more concentrated on the subject area in the training-based version.

\subsection{SubjectSpatial200K dataset}
\begin{figure}[t]
    \centering
    \includegraphics[width=\linewidth]{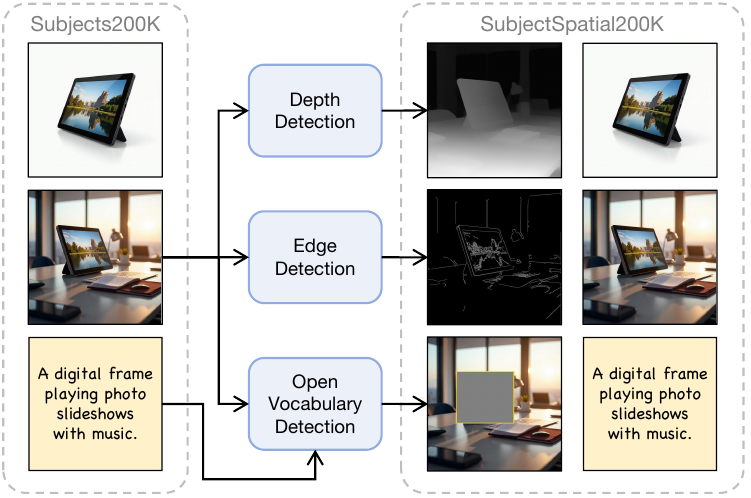}
    \vspace{-1.5em}
    \caption{SubjectSpatial200K dataset construction pipeline.}
    \label{fig:data_pipeline}
    \vspace{-1em}
\end{figure}

Our SubjectSpatial200K dataset aims to address the lack of a publicly available dataset for multi-conditional generative tasks. Existing datasets fail to include both the subject-driven and spatially-aligned annotations. Recently, the Subjects200K \cite{ominicontrol} dataset provides a publicly accessible dataset for subject-driven generation. Based on it, we introduce the SubjectSpatial200K dataset, which is a unified high-quality dataset designed for training and testing multi-conditional controllable generative models. This dataset includes comprehensive annotations as elaborated below. Besides, the construction pipeline is detailed in \cref{fig:data_pipeline}.

\noindent \textbf{Subject Grounding Annotation.}
 The subject grounding annotation is significantly necessary for many generative tasks like instance-level inpainting \cite{powerpaint, brushnet}, instance-level controllable generation \cite{gligen, instancediffusion}, and object insertion \cite{objectStitch, anydoor}. 
 By leveraging the open-vocabulary object detection model Mamba-YOLO-World \cite{mambayoloworld} on Subjects200K, we detect bounding boxes for all subjects according to their category descriptions and subsequently derive the corresponding mask regions.
 
\noindent \textbf{Spatial Map Annotation.}
The spatial map annotation further extends the applicable scope of our dataset to spatially-aligned synthesis tasks. Specifically, we employ the Depth-Anything \cite{depthanything} model and the OpenCV \cite{opencv_library} library on Subjects200K to derive the Depth and Canny maps. 

%% file: sec/4_experiment.tex
\input{sec/table}

\section{Experiment}
\label{sec:exp}

\subsection{Setup}

\noindent \textbf{Implementation.}
We use the FLUX.1-schnell \cite{flux} as our base model and the weights provided by OminiControl \cite{ominicontrol} as our pre-trained Condition-LoRA module weights.
During the training of our Denoising-LoRA module, we use a rank of 4, consistent with the Condition-LoRA. We choose the Adam optimizer with a learning rate of $1e^{-4}$ and set the weight decay to 0.01. Our models are trained for 30,000 steps on 16 NVIDIA V100 GPUs at a resolution of $512\times512$.

\begin{figure}[t]
    \centering
    \includegraphics[width=1\linewidth]{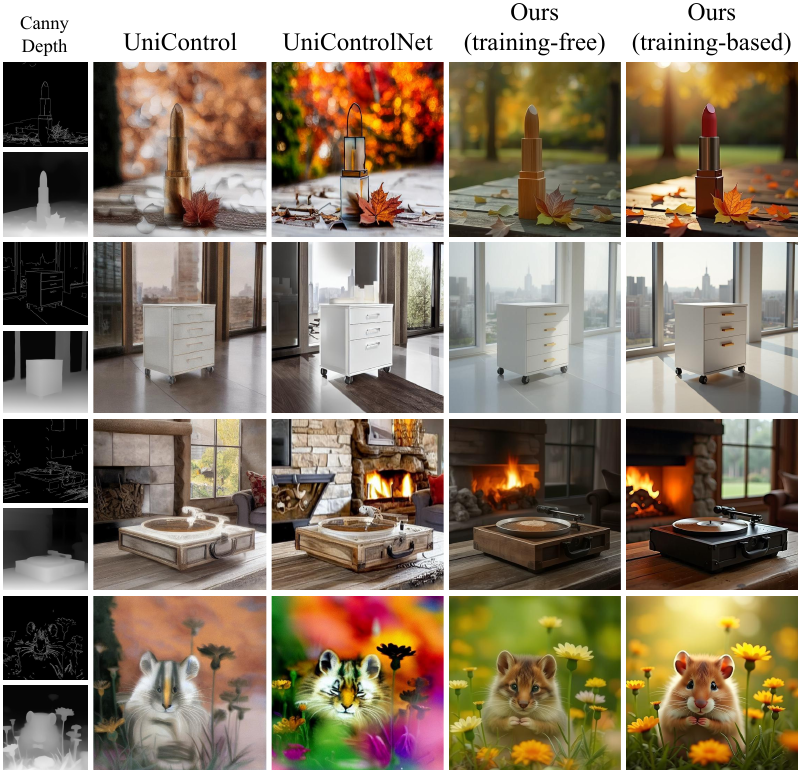}
    \vspace{-1.5em}
    \caption{Qualitative comparison on Multi-Spatial generation.}
    \label{fig:multi_spatial}
    \vspace{-1em}
\end{figure}

\noindent \textbf{Benchmarks.}
We evaluate the performance of our method in both training-free and training-based versions. The training and testing datasets are partitioned from the SubjectSpatial200K dataset based on image quality assessment scores evaluated by ChatGPT-4o, with details provided in \cref{sec:sup_dataset}. Importantly, the dataset partitioning scheme remains consistent in all experiments.

\begin{figure}[t]
    \centering
    \includegraphics[width=1\linewidth]{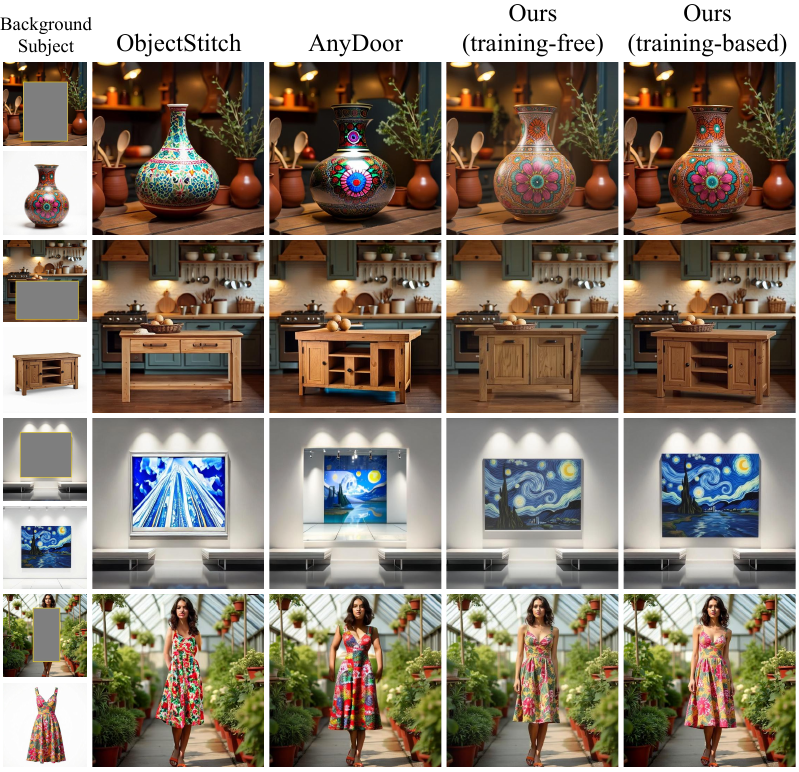}
    \vspace{-1.5em}
    \caption{Qualitative comparison on Subject-Insertion generation.}
    \label{fig:insertion}
    \vspace{-1em}
\end{figure}

\noindent \textbf{Metrics.}
To evaluate the subject consistency, we calculate the CLIP-I \cite{clip} score and DINO \cite{dino} score between the generated images and the ground truth images.
To assess the generative quality, we compute the FID \cite{fid} and SSIM \cite{ssim} between the generated image set and the ground truth image set.
To measure the controllability, we compute the F1 Score for edge conditions and the MSE score for depth conditions between the extracted maps from generated images and the original conditions. 
Additionally, we adopt the CLIP-T \cite{clip} score to estimate the text consistency between the generated images and the text prompts.

\subsection{Main Result}
We conduct extensive and comprehensive comparative experiments on the Multi-Spatial, Subject-Insertion, and Subject-Spatial conditional generative tasks.

\subsubsection{Multi-Spatial Conditional Generation}
The Multi-Spatial conditional generation aims to generate images adhering to the collective layout constraints of diverse spatial conditions. This requires the model to achieve a more comprehensive layout control based on input conditions in a complementary manner. The comparative results in \cref{tab:main_results} and \cref{fig:multi_spatial} demonstrate that our method outperforms existing multi-spatial conditional generation approaches in generative quality and controllability.

\begin{figure}[t]
    \centering
    \includegraphics[width=1\linewidth]{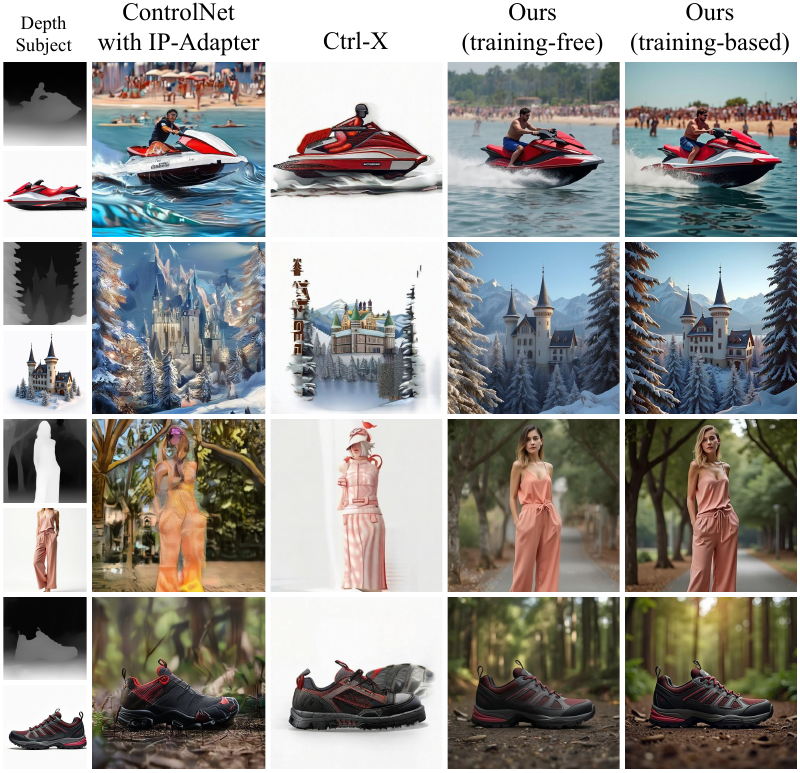}
    \vspace{-1.5em}
    \caption{Qualitative comparison on Subject-Depth generation.}
    \label{fig:subject_depth}
    \vspace{-1em}
\end{figure}

\subsubsection{Subject-Insertion Conditional Generation}
The Subject-Insertion conditional generation requires the model to generate images where the reference subject is inserted into the masked region of the target background.  As illustrated in \cref{tab:main_results} and \cref{fig:insertion}, our UniCombine demonstrates superior performance compared to previous methods with three advantages: \textbf{Firstly}, our method ensures that the reference subject is inserted into the background with high consistency and harmonious integration. \textbf{Secondly,} our method excels in open-world object insertion without requiring test-time tuning, unlike conventional customization methods \cite{dreambooth, customDiffusion}. \textbf{Finally}, our method demonstrates strong semantic comprehension capabilities, enabling it to extract the desired object from a complex subject image with a non-white background, rather than simply pasting the entire subject image into the masked region.

\subsubsection{Subject-Spatial Conditional Generation}
\begin{figure}[t]
    \centering
    \includegraphics[width=1\linewidth]{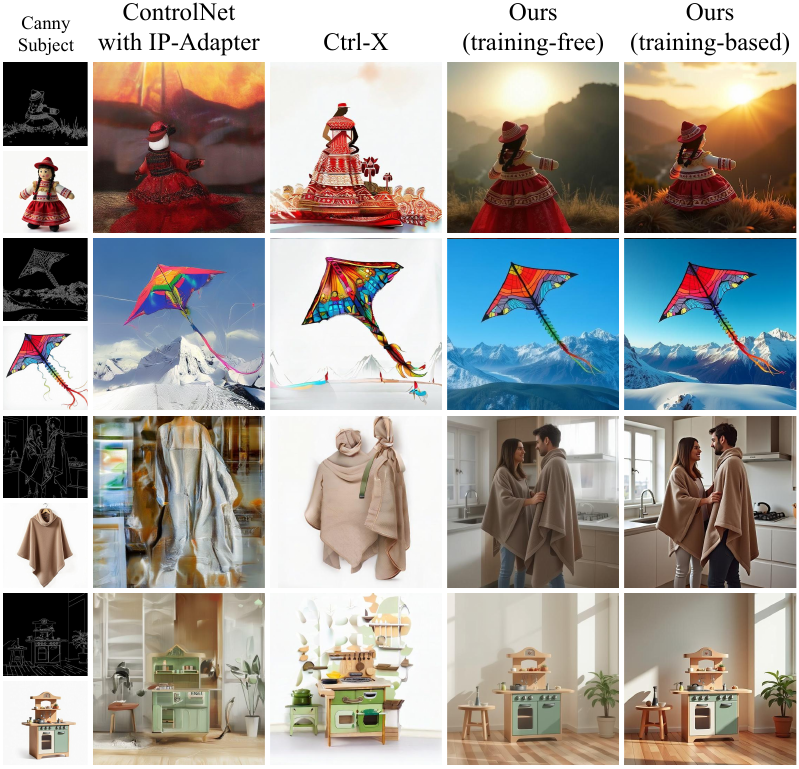}
    \vspace{-1.5em}
    \caption{Qualitative comparison on Subject-Canny generation.}
    \label{fig:subject_canny}
    \vspace{-1em}
\end{figure}

The Subject-Spatial conditional generation focuses on generating images of the reference subject while ensuring the layout aligns with specified spatial conditions. We compare our method with Ctrl-X \cite{ctrlx} and a simple baseline model. Ctrl-X is a recently proposed model based on SDXL \cite{sdxl} that simultaneously controls structure and appearance. The baseline model is constructed by integrating the FLUX ControlNet \cite{XLabs-AI_Flux-ControlNet-Canny, XLabs-AI_Flux-ControlNet-Depth} and FLUX IP-Adapter \cite{XLabs-AI_Flux-ip-adapter} into the FLUX.1-dev \cite{flux} base model.
Specifically, we divided the Subject-Spatial generative task into different experimental groups based on the type of spatial conditions, referred to as Subject-Depth and Subject-Canny, respectively. 
As presented in \cref{fig:subject_depth}, \cref{fig:subject_canny}, and \cref{tab:main_results}, the experimental results demonstrate the superior performance of our UniCombine: \textbf{Firstly}, our method exhibits stronger semantic comprehension capability, generating the reference subject in the accurate localization of the spatial conditions without confusing appearance features. \textbf{Secondly}, our method demonstrates greater adaptability, generating the reference subject with reasonable morphological transformations to align with the guidance of spatial conditions and text prompts. \textbf{Lastly}, our method achieves superior subject consistency while maintaining excellent spatial coherence.

\subsubsection{Textual Guidance}
As shown in \cref{fig:first_image} and \cref{tab:main_results}, our method not only allows for controllable generation by combining multiple conditions but also enables precise textual guidance simultaneously. By utilizing a unified input sequence \(S= [T; X; C_{1}; \dots; C_{N}]\) during the denoising process, our UniCombine effectively aligns the descriptive words in $T$ with the relevant features in $C_i$ and the corresponding patches in $X$, thereby achieving a remarkable text-guided multi-conditional controllable generation.

\subsection{Ablation Study}
We exhibit the ablation study results conducted on the Subject-Insertion task in this section, while more results on the other tasks are provided in \cref{sec:sup_cmmdit}.

\begin{figure}[t]
    \centering
    \includegraphics[width=\linewidth]{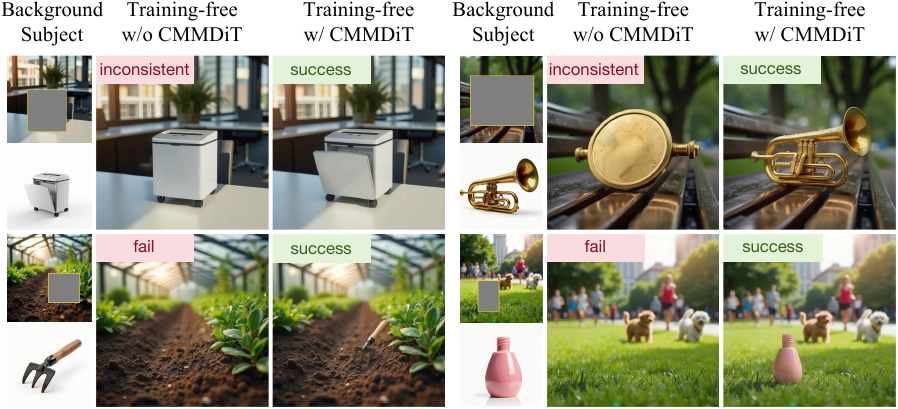}
    \vspace{-2em}
    \caption{Qualitative ablation of CMMDiT Attention mechanism on training-free Subject-Insertion task.}
    \label{fig:cmmdit}
\end{figure}

\noindent \textbf{Effect of Conditional MMDiT Attention.}
To evaluate the effectiveness of our proposed Conditional MMDiT Attention mechanism, we replace the CMMDiT Attention with the original MMDiT Attention and test its training-free performance to avoid the influence of training data. 
As shown in \cref{tab:cmmdit} and \cref{fig:cmmdit}, our framework attains superior performance with fewer attention operations when employing the CMMDiT Attention mechanism.

\begin{table}[t]
  \footnotesize
  \centering
    \begin{tabular}{lcccc}
    \toprule
        Method & CLIP-I $\uparrow$ & DINO $\uparrow$ & CLIP-T $\uparrow$  & AttnOps $\downarrow$ \\
    \midrule
        Ours w/o CMMDiT & 95.47 & 88.42 & 33.10 & 732.17M\\
        Ours w/ CMMDiT & \textbf{95.60} & \textbf{89.01}& \textbf{33.11} & \textbf{612.63M}\\ 
    \bottomrule
    \end{tabular}
    \vspace{-1em}
    \caption{Quantitative ablation of CMMDiT Attention mechanism on training-free Subject-Insertion task. AttnOps is short for the number of attention operations.}
    \label{tab:cmmdit}
\end{table}

\begin{table}[t]
    \footnotesize
    \centering
    \begin{tabular}{lccc}
    \toprule
        Method & CLIP-I $\uparrow$ & DINO $\uparrow$ & CLIP-T $\uparrow$ \\
    \midrule
        Ours w/ Text-LoRA & 96.97 & 92.32 & \textbf{33.10} \\
        Ours w/ Denoising-LoRA  & \textbf{97.14} & \textbf{92.96} & 33.08 \\ 
    \bottomrule
    \end{tabular}
    \vspace{-1em}
    \caption{Quantitative ablation of trainable LoRA on training-based Subject-Insertion task.}
    \label{tab:lora}
\end{table}

\begin{figure}[t]
    \centering
    \includegraphics[width=\linewidth]{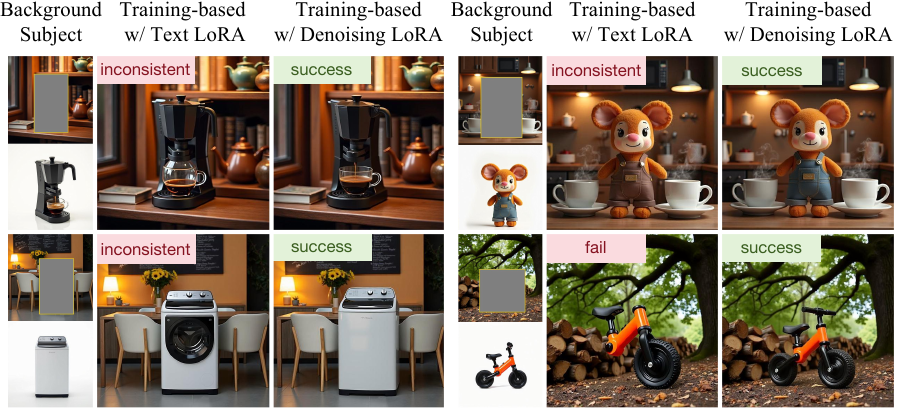}
    \vspace{-2em}
    \caption{Qualitative ablation of trainable LoRA on training-based Subject-Insertion task.}
    \label{fig:lora}
    \vspace{-1em}

\end{figure}

\noindent \textbf{Different Options for Trainable LoRA.}
To evaluate whether the trainable LoRA module can be applied to the text branch instead of the denoising branch, we load a Text-LoRA in the text branch, with a configuration identical to that of the Denoising-LoRA. The \cref{tab:lora} and \cref{fig:lora} indicate that applying the trainable LoRA module to the denoising branch better modulates the feature aggregation operation across multiple conditional branches.

\begin{figure}[t]
    \centering
    \includegraphics[width=\linewidth]{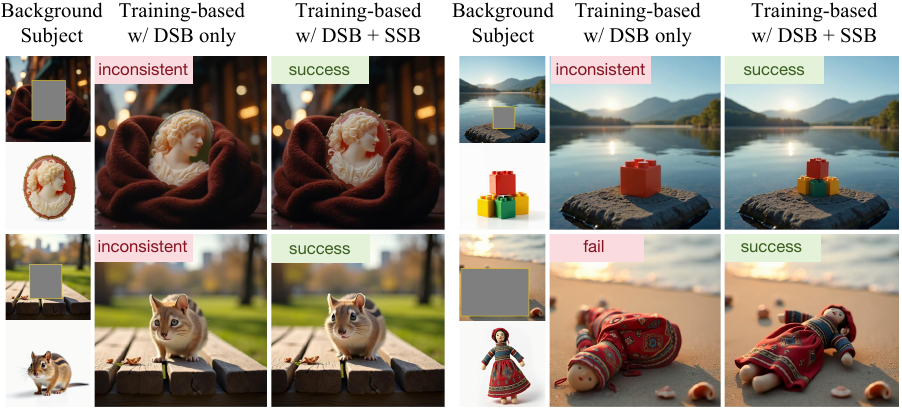}
    \vspace{-2em}
    \caption{Qualitative ablation of training strategy on training-based Subject-Insertion task. DSB: Dual-Stream Blocks. SSB: Single-Stream Blocks.}
    \label{fig:half}
\end{figure}

\noindent \textbf{Training Strategy.}
As the parameter scale of the base model increases, the FLUX adaptations of ControlNet \cite{XLabs-AI_Flux-ControlNet-Canny, XLabs-AI_Flux-ControlNet-Depth} and IP-adapter \cite{XLabs-AI_Flux-ip-adapter} provided by the HuggingFace \cite{diffusers} community inject conditional features only into the dual-stream MMDiT blocks, rather than the entire network, to save memory. In contrast, since our Denoising-LoRA module introduces only a small number of parameters, we incorporate it into both the dual-stream and single-stream blocks to achieve better performance. The results in \cref{tab:half} and \cref{fig:half} confirm the validity of our choice.

\begin{table}[t]
    \footnotesize
    \centering
    \begin{tabular}{lccc}
    \toprule
        Method & CLIP-I $\uparrow$ & DINO $\uparrow$ & CLIP-T $\uparrow$ \\
    \midrule
        Ours w/ DSB only  & 96.85 & 92.38 & 33.07\\
        Ours w/ DSB and SSB  & \textbf{97.14} & \textbf{92.96} & \textbf{33.08} \\ 
    \bottomrule
    \end{tabular}
    \vspace{-1em}
    \caption{Quantitative ablation of training strategy on training-based Subject-Insertion task. DSB: Dual-Stream Blocks. SSB: Single-Stream Blocks.}
    \label{tab:half}
\end{table}

\begin{table}[t]
    \footnotesize
    \centering
    \begin{tabular}{lcc}
    \toprule
        Model & GPU Memory $\downarrow$ & Add Params $\downarrow$ \\
    \midrule
        FLUX (bf16, base model) & 32933M & -\\ 
    \midrule
        CN, 1 cond & 35235M & 744M\\
        IP, 1 cond & 35325M & 918M\\
        CN + IP, 2 cond & 36753M & 1662M\\
        Ours (training-free), 2 cond & \textbf{33323M} & \textbf{29M} \\ 
        Ours (training-based), 2 cond &  \underline{33349M} &  \underline{44M} \\ 
    \bottomrule
    \end{tabular}
    \vspace{-1em}
    \caption{Comparison of inference GPU memory cost and additionally introduced parameters. CN: ControlNet. IP: IP-Adapter.}
    \label{tab:memory}
    \vspace{-1em}
\end{table}

\noindent \textbf{Computational Cost.}
The overheads of our approach in terms of inference GPU memory cost and additionally introduced parameters are minimal. The comparison results against the FLUX ControlNet \cite{XLabs-AI_Flux-ControlNet-Depth, XLabs-AI_Flux-ControlNet-Canny} and FLUX IP-Adapter \cite{XLabs-AI_Flux-ip-adapter} are shown in \cref{tab:memory}.

\begin{figure}[ht]
    \centering
    \includegraphics[width=\linewidth]{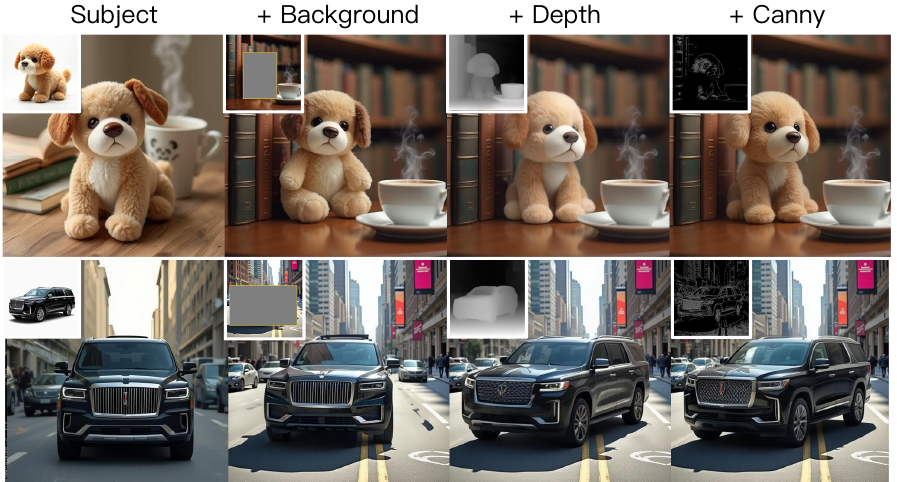}
    \vspace{-2em}
    \caption{From left to right are training-free multi-conditional combination tasks under: 1/2/3/4 conditions.}
    \label{fig:ab_multi_cond}

\end{figure}

\noindent \textbf{More Conditional Branches.}
Our model places no restrictions on the number of supported conditions. The results shown in  \cref{fig:ab_multi_cond} demonstrate our model’s strong scalability. As the number of conditional branches increases, the level of control becomes finer.

\begin{figure}[ht]
    \centering
    \includegraphics[width=\linewidth]{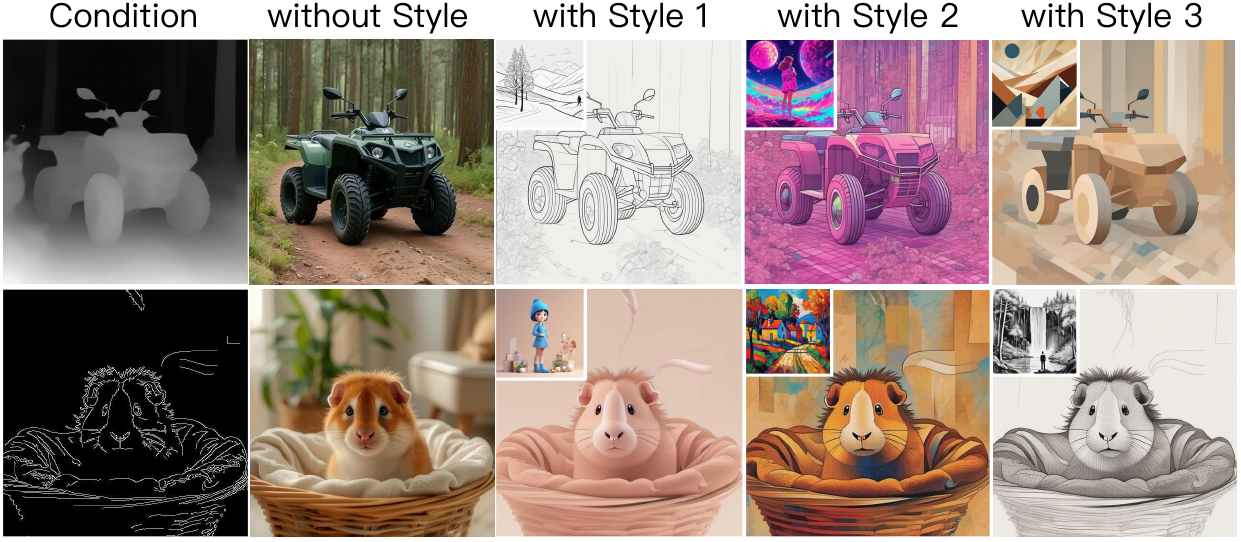}
    \vspace{-2em}
    \caption{Training-free Spatial-Style combination task.}
    \label{fig:ab_style}
\end{figure}

\noindent \textbf{More Application Scenarios.}
Our UniCombine can be easily extended to new scenarios, such as reference-based image stylization. After training a new Condition-LoRA on StyleBooth \cite{stylebooth} dataset, our UniCombine is able to integrate the style of the reference image with other conditions successfully, as demonstrated in \cref{fig:ab_style}.

%% file: sec/table.tex
\begin{table*}[t] 
\footnotesize
\centering
\begin{tabular}{c|l|cc|cc|cc|c}
\toprule

\multirow{2}{*}{Task} & \multirow{2}{*}{Method} & \multicolumn{2}{c|}{Generative Quality} & \multicolumn{2}{c|} {Controllability}  & \multicolumn{2}{c|}{Subject Consistency} & Text Consistency \\
& & $\text{FID}\downarrow$ & $\text{SSIM}\uparrow$ & $\text{F1}\uparrow $ & $\text{MSE}\downarrow$ & $\text{CLIP-I}\uparrow$ & $\text{DINO}\uparrow$ & $\text{CLIP-T}\uparrow$  \\
\midrule

\multirow{4}{*}{Multi-Spatial} & UniControl & 44.17 & 0.32 & 0.07 & 1346.02 & - & - & 30.28\\
 & UniControlNet  & 20.96 & 0.28 & 0.09 & 1231.06 & - & - & 32.74 \\
 & UniCombine (training-free)  & \underline{10.35} & \underline{0.54} & \underline{0.18} & \underline{519.53} & - & - & \textbf{33.70}\\
 & UniCombine (training-based)  & \textbf{6.82} & \textbf{0.64} & \textbf{0.24} & \textbf{165.90} & - & - & \underline{33.45}\\
\midrule

\multirow{4}{*}{Subject-Insertion} & ObjectStitch & 26.86 & 0.37 & - & - & 93.05 & 82.34 & 32.25\\
 & AnyDoor  & 26.07 & 0.37 & - & - & 94.88 & 86.04 & 32.55\\
 & UniCombine (training-free)  & \underline{6.37} & \underline{0.76} & - &- & \underline{95.60} & \underline{89.01} & \textbf{33.11} \\
 & UniCombine (training-based)  & \textbf{4.55} & \textbf{0.81} & - & - & \textbf{97.14} & \textbf{92.96} & \underline{33.08}\\
\midrule

\multirow{4}{*}{Subject-Depth} & ControlNet w. IP-Adapter & 29.93 & 0.34 & - & 1295.80 & 80.41 & 62.26 & 32.94\\
 & Ctrl-X & 52.37 & 0.36 & - & 2644.90 & 78.08 & 50.83 & 30.20\\
 & UniCombine (training-free)  & \underline{10.03} & \underline{0.48} & - & \underline{507.40} & \underline{91.15} & \underline{85.73} & \textbf{33.41}\\
 & UniCombine (training-based)  & \textbf{6.66} & \textbf{0.55} & - & \textbf{196.65} & \textbf{94.47} & \textbf{90.31} & \underline{33.30}\\
\midrule

\multirow{4}{*}{Subject-Canny} & ControlNet w. IP-Adapter & 30.38 & 0.38 & 0.09 & - & 79.80 & 60.19 & 32.85\\
 & Ctrl-X & 47.89 & 0.36 & 0.05 & - & 79.35 & 54.31 & 30.34\\
 & UniCombine (training-free)  & \underline{10.22} & \underline{0.49} & \underline{0.17} & - & \underline{91.84} & \underline{86.88} & \underline{33.21} \\
 & UniCombine (training-based) & \textbf{6.01} & \textbf{0.61} & \textbf{0.24} & - & \textbf{95.26} & \textbf{92.59} & \textbf{33.30} \\

\bottomrule
\end{tabular}
\vspace{-0.5em}
\caption{Quantitative comparison of our method with existing approaches on Multi-Spatial, Subject-Insertion, Subject-Depth, and Subject-Canny conditional generative tasks. The \textbf{bold} and \underline{underlined} figures represent the optimal and sub-optimal results, respectively.}
\label{tab:main_results}
\vspace{-1em}

\end{table*}

%% file: sec/5_conclusion.tex
\section{Conclusion}
We present UniCombine, a DiT-based multi-conditional controllable generative framework capable of handling any combination of conditions, including but not limited to text prompts, spatial maps, and subject images. Extensive experiments
on Subject-Insertion, Subject-Spatial, and Multi-Spatial 
conditional generative tasks demonstrate the state-of-the-art performance of our UniCombine 
in both training-free and training-based versions. 
Additionally, we propose the SubjectSpatial200K dataset to address the lack of a publicly available dataset for training and testing multi-conditional generative models. We believe our work can advance the development of the controllable generation field.


%% file: sec/6_supplement.tex
\clearpage
\renewcommand{\thefigure}{A\arabic{figure}}
\setcounter{figure}{0}
\renewcommand{\thetable}{A\arabic{table}}
\setcounter{table}{0}
\renewcommand{\thesection}{A\arabic{section}}
\setcounter{section}{0}
\maketitlesupplementary

\section{Dataset Partitioning Scheme}
\label{sec:sup_dataset}
In our proposed SubjectSpatial200K dataset, we utilize the ChatGPT-4o assessment scores provided by Subjects200K \cite{ominicontrol} on Subject Consistency, Composition Structure, and Image Quality to guide the dataset partitioning in our experiments.
\begin{itemize}
    \item Subject Consistency: Ensuring the identity of the subject image is consistent with that of the ground truth image.
    \item Composition Structure: Verifying a reasonable composition of the subject and ground truth images.
    \item Image Quality: Confirming each image pair maintains high resolution and visual fidelity.
\end{itemize}
We partition the dataset into 139,403 training samples and 5,827 testing samples through \cref{algo:dataset}.

\vspace{-1em}
\begin{algorithm}
\caption{Dataset Partitioning Scheme}
\label{algo:dataset}
\KwIn{example}
\KwOut{train or test}
cs $\gets$ example[``Composite Structure'']\\
iq $\gets$ example[``Image Quality'']\\
sc $\gets$ example[``Subject Consistency'']\\
scores $\gets$ [cs, iq, sc]\\
\If{ \textbf{all}($s == 5$ \textbf{for} $s$ \textbf{in} $scores$) }{
    \Return train\;
}
\ElseIf{$cs \geq 3$ \textbf{and} $iq == 5$ \textbf{and} $sc == 5$ }{
    \Return test\;
}
\end{algorithm}
\vspace{-1em}

\section{More Ablation on CMMDiT Attention}
\label{sec:sup_cmmdit}
\begin{figure}[ht]
    \centering
    \includegraphics[width=\linewidth]{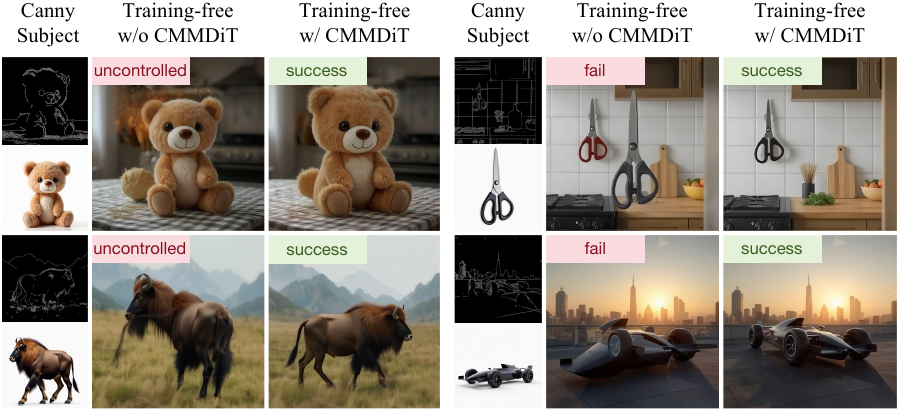}
    \vspace{-2em}
    \caption{Qualitative ablation of CMMDiT Attention mechanism on training-free Subject-Canny task}
    \vspace{-0.5em}
    \label{fig:cmmdit_spp1}
\end{figure}

\begin{figure}[ht]
    \centering
    \includegraphics[width=\linewidth]{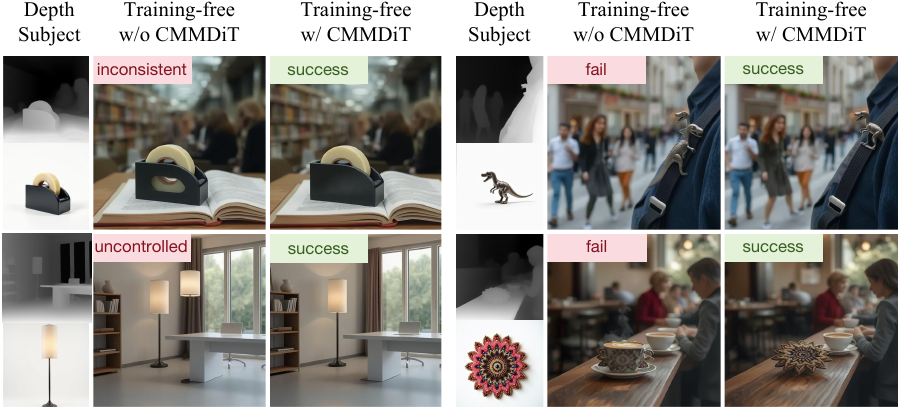}
    \vspace{-2em}
    \caption{Qualitative ablation of CMMDiT Attention mechanism on training-free Subject-Depth task}
    \label{fig:cmmdit_spp2}
    \vspace{-0.5em}
\end{figure}

\begin{figure}[ht]
    \centering
    \includegraphics[width=\linewidth]{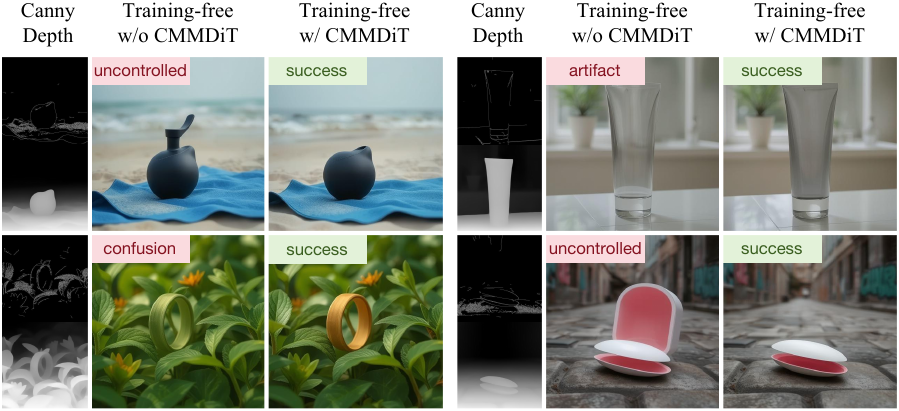}
    \vspace{-2em}
    \caption{Qualitative ablation of CMMDiT Attention mechanism on training-free Multi-Spatial task}
    \label{fig:cmmdit_spp3}
    \vspace{-1em}
\end{figure}

 More quantitative and qualitative ablation results on the other multi-conditional generative tasks are provided here. The comprehensive ablation results in \cref{tab:cmmdit_spp1}, \cref{tab:cmmdit_spp2}, \cref{tab:cmmdit_spp3}, \cref{fig:cmmdit_spp1}, \cref{fig:cmmdit_spp2}, and \cref{fig:cmmdit_spp3} demonstrate that the UniCombine performs better with our proposed CMMDiT Attention.

\begin{table}[ht]
    \footnotesize
    \centering
    \begin{tabular}{lcccc}
    \toprule
        Method & CLIP-I $\uparrow$ & DINO $\uparrow$ & CLIP-T $\uparrow$ & F1 $\uparrow$ \\
    \midrule
        Ours w/o CMMDiT & 91.51 & 86.31 & 33.20 & 0.16\\
    \midrule
        Ours w/ CMMDiT  & \textbf{91.84} & \textbf{86.88} & \textbf{33.21}  & \textbf{0.17}\\ 
    \bottomrule
    \end{tabular}
    \vspace{-1em}
    \caption{Quantitative ablation of CMMDiT Attention mechanism on training-free Subject-Canny task}
    \label{tab:cmmdit_spp1}
    \vspace{-1em}
\end{table}

\section{More Qualitative Results}
\begin{figure*}[t]
    \centering
    \includegraphics[width=\textwidth]{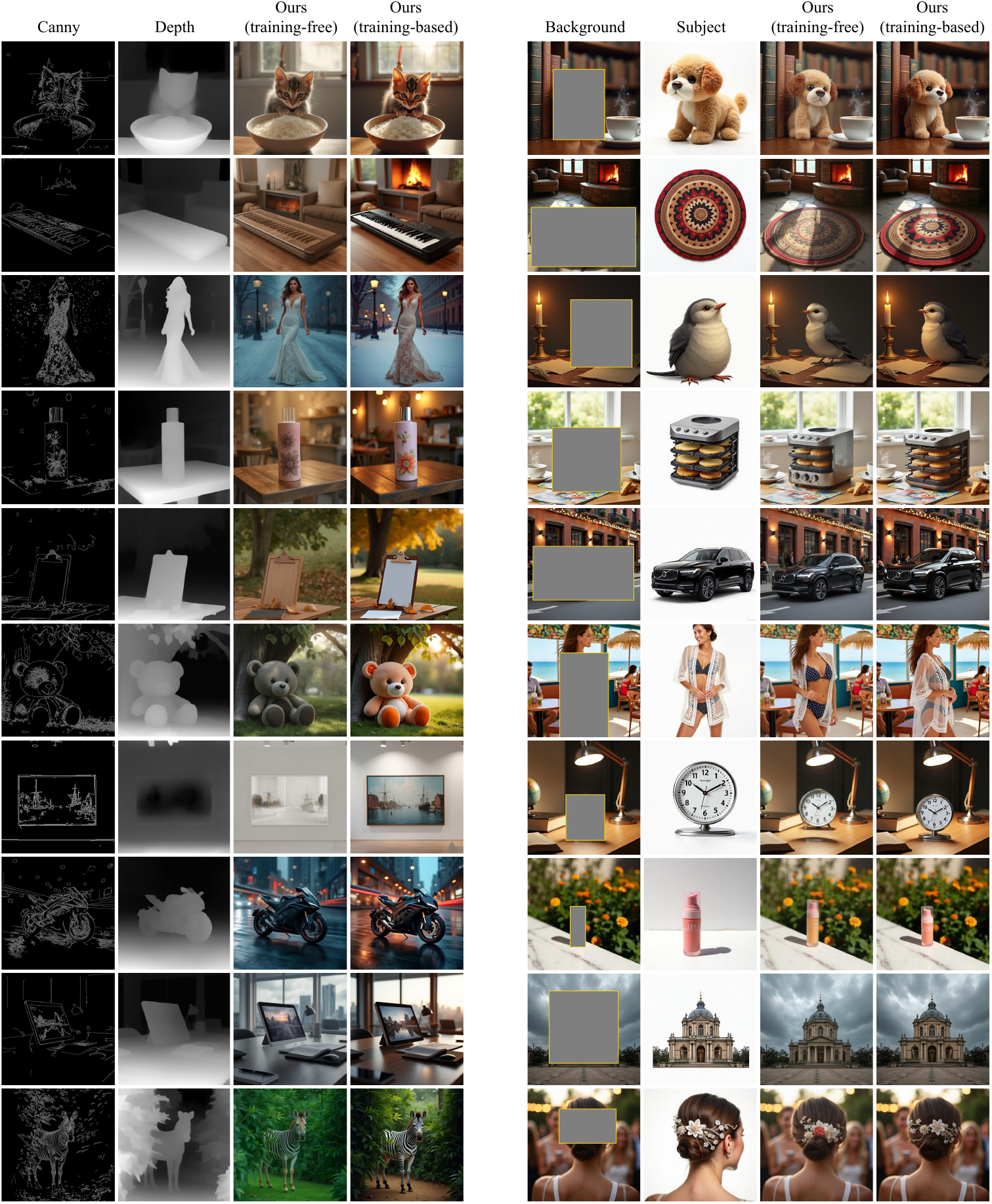}
    \vspace{-1em}
    \caption{More qualitative results on Multi-Spatial and Subject-Insertion tasks.}
    \label{fig:sup1}
\end{figure*}

\begin{figure*}[t]
    \centering
    \includegraphics[width=\textwidth]{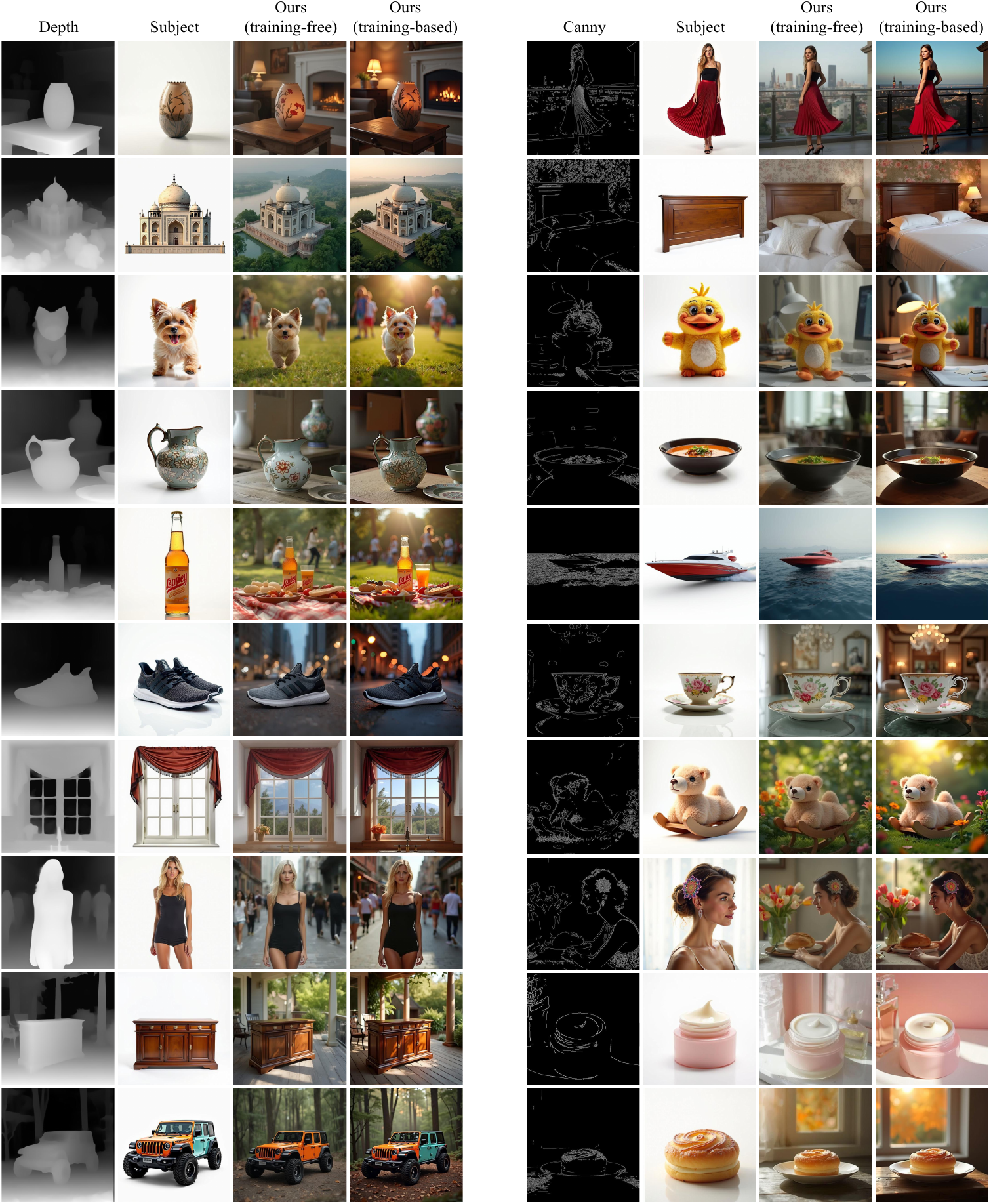}
    \vspace{-1em}
    \caption{More qualitative results on Subject-Depth and Subject-Canny tasks.}
    \label{fig:sup2}
\end{figure*}

More qualitative results are presented in \cref{fig:sup1} and \cref{fig:sup2}.

\begin{table}[ht]
    \footnotesize
    \centering
    \begin{tabular}{lcccc}
    \toprule
        Method & CLIP-I $\uparrow$ & DINO $\uparrow$ & CLIP-T $\uparrow$ & MSE $\downarrow$  \\
    \midrule
        \makecell[l]{Ours w/o CMMDiT} & 90.83 & 85.38 & 33.38 & 547.63 \\
    \midrule
        \makecell[l]{Ours w/ CMMDiT}  & \textbf{91.15} & \textbf{85.73} & \textbf{33.41} &\textbf{507.40} \\ 
    \bottomrule
    \end{tabular}
    \vspace{-1em}
    \caption{Quantitative ablation of CMMDiT Attention mechanism on training-free Subject-Depth task}
    \label{tab:cmmdit_spp2}
    \vspace{-1em}
\end{table}

\begin{table}[ht]
    \footnotesize
    \centering
    \begin{tabular}{lcccc}
    \toprule
        Method & CLIP-T $\uparrow$ & F1 $\uparrow$ & MSE $\downarrow$  \\
    \midrule
        \makecell[l]{Ours w/o CMMDiT} & 33.70 & 0.17 & 524.04   \\
    \midrule
        \makecell[l]{Ours w/ CMMDiT}  & 33.70  & \textbf{0.18} & \textbf{519.53} \\ 
    \bottomrule
    \end{tabular}
    \vspace{-1em}
    \caption{Quantitative ablation of CMMDiT Attention mechanism on training-free Multi-Spatial task}
    \label{tab:cmmdit_spp3}
    \vspace{-1em}
\end{table}

%% file: main.bbl
\begin{thebibliography}{61}
\providecommand{\natexlab}[1]{#1}
\providecommand{\url}[1]{\texttt{#1}}
\expandafter\ifx\csname urlstyle\endcsname\relax
  \providecommand{\doi}[1]{doi: #1}\else
  \providecommand{\doi}{doi: \begingroup \urlstyle{rm}\Url}\fi

\bibitem[Bradski(2000)]{opencv_library}
G. Bradski.
\newblock {The OpenCV Library}.
\newblock \emph{Dr. Dobb's Journal of Software Tools}, 2000.

\bibitem[Caron et~al.(2021)Caron, Touvron, Misra, J{\'e}gou, Mairal, Bojanowski, and Joulin]{dino}
Mathilde Caron, Hugo Touvron, Ishan Misra, Herv{\'e} J{\'e}gou, Julien Mairal, Piotr Bojanowski, and Armand Joulin.
\newblock Emerging properties in self-supervised vision transformers.
\newblock In \emph{Proceedings of the IEEE/CVF international conference on computer vision}, pages 9650--9660, 2021.

\bibitem[Chen et~al.(2024{\natexlab{a}})Chen, Zhang, He, Peng, and Niu]{chen2024mureobjectstitch}
Jiaxuan Chen, Bo Zhang, Qingdong He, Jinlong Peng, and Li Niu.
\newblock Mureobjectstitch: Multi-reference image composition.
\newblock \emph{arXiv preprint arXiv:2411.07462}, 2024{\natexlab{a}}.

\bibitem[Chen et~al.(2024{\natexlab{b}})Chen, Huang, Liu, Shen, Zhao, and Zhao]{anydoor}
Xi Chen, Lianghua Huang, Yu Liu, Yujun Shen, Deli Zhao, and Hengshuang Zhao.
\newblock Anydoor: Zero-shot object-level image customization.
\newblock In \emph{Proceedings of the IEEE/CVF Conference on Computer Vision and Pattern Recognition}, pages 6593--6602, 2024{\natexlab{b}}.

\bibitem[Chong et~al.(2024)Chong, Dong, Li, Zhang, Zhang, Zhang, Zhao, Jiang, and Liang]{catvton}
Zheng Chong, Xiao Dong, Haoxiang Li, Shiyue Zhang, Wenqing Zhang, Xujie Zhang, Hanqing Zhao, Dongmei Jiang, and Xiaodan Liang.
\newblock Catvton: Concatenation is all you need for virtual try-on with diffusion models.
\newblock \emph{arXiv preprint arXiv:2407.15886}, 2024.

\bibitem[Couairon et~al.(2023)Couairon, Careil, Cord, Lathuiliere, and Verbeek]{ZestGuide}
Guillaume Couairon, Marlene Careil, Matthieu Cord, St{\'e}phane Lathuiliere, and Jakob Verbeek.
\newblock Zero-shot spatial layout conditioning for text-to-image diffusion models.
\newblock In \emph{Proceedings of the IEEE/CVF International Conference on Computer Vision}, pages 2174--2183, 2023.

\bibitem[Esser et~al.()Esser, Kulal, Blattmann, Entezari, M{\"u}ller, Saini, Levi, Lorenz, Sauer, Boesel, et~al.]{sd3}
Patrick Esser, Sumith Kulal, Andreas Blattmann, Rahim Entezari, Jonas M{\"u}ller, Harry Saini, Yam Levi, Dominik Lorenz, Axel Sauer, Frederic Boesel, et~al.
\newblock Scaling rectified flow transformers for high-resolution image synthesis, 2024.
\newblock \emph{URL https://arxiv. org/abs/2403.03206}, 2.

\bibitem[Gal et~al.(2022)Gal, Alaluf, Atzmon, Patashnik, Bermano, Chechik, and Cohen-Or]{textualInverion}
Rinon Gal, Yuval Alaluf, Yuval Atzmon, Or Patashnik, Amit~H Bermano, Gal Chechik, and Daniel Cohen-Or.
\newblock An image is worth one word: Personalizing text-to-image generation using textual inversion.
\newblock \emph{arXiv preprint arXiv:2208.01618}, 2022.

\bibitem[Goodfellow et~al.(2020)Goodfellow, Pouget-Abadie, Mirza, Xu, Warde-Farley, Ozair, Courville, and Bengio]{gan}
Ian Goodfellow, Jean Pouget-Abadie, Mehdi Mirza, Bing Xu, David Warde-Farley, Sherjil Ozair, Aaron Courville, and Yoshua Bengio.
\newblock Generative adversarial networks.
\newblock \emph{Communications of the ACM}, 63\penalty0 (11):\penalty0 139--144, 2020.

\bibitem[Han et~al.(2024)Han, Mao, Jiang, Pan, and Zhang]{stylebooth}
Zhen Han, Chaojie Mao, Zeyinzi Jiang, Yulin Pan, and Jingfeng Zhang.
\newblock Stylebooth: Image style editing with multimodal instruction.
\newblock \emph{arXiv preprint arXiv:2404.12154}, 2024.

\bibitem[Hertz et~al.(2022)Hertz, Mokady, Tenenbaum, Aberman, Pritch, and Cohen-Or]{prompt2promt}
Amir Hertz, Ron Mokady, Jay Tenenbaum, Kfir Aberman, Yael Pritch, and Daniel Cohen-Or.
\newblock Prompt-to-prompt image editing with cross attention control.
\newblock \emph{arXiv preprint arXiv:2208.01626}, 2022.

\bibitem[Heusel et~al.(2017)Heusel, Ramsauer, Unterthiner, Nessler, and Hochreiter]{fid}
Martin Heusel, Hubert Ramsauer, Thomas Unterthiner, Bernhard Nessler, and Sepp Hochreiter.
\newblock Gans trained by a two time-scale update rule converge to a local nash equilibrium.
\newblock \emph{Advances in neural information processing systems}, 30, 2017.

\bibitem[Ho et~al.(2020)Ho, Jain, and Abbeel]{ddpm}
Jonathan Ho, Ajay Jain, and Pieter Abbeel.
\newblock Denoising diffusion probabilistic models.
\newblock \emph{Advances in neural information processing systems}, 33:\penalty0 6840--6851, 2020.

\bibitem[Hu et~al.(2023{\natexlab{a}})Hu, Zheng, Liu, Zheng, Wang, Tao, and Cham]{cocktail}
Minghui Hu, Jianbin Zheng, Daqing Liu, Chuanxia Zheng, Chaoyue Wang, Dacheng Tao, and Tat-Jen Cham.
\newblock Cocktail: Mixing multi-modality control for text-conditional image generation.
\newblock In \emph{Thirty-seventh Conference on Neural Information Processing Systems}, 2023{\natexlab{a}}.

\bibitem[Hu et~al.(2023{\natexlab{b}})Hu, Yi, Zhu, Liu, Peng, Wang, Wang, and Ma]{hu2023stroke}
Teng Hu, Ran Yi, Haokun Zhu, Liang Liu, Jinlong Peng, Yabiao Wang, Chengjie Wang, and Lizhuang Ma.
\newblock Stroke-based neural painting and stylization with dynamically predicted painting region.
\newblock In \emph{Proceedings of the 31st ACM International Conference on Multimedia}, pages 7470--7480, 2023{\natexlab{b}}.

\bibitem[HuggingFace(2023)]{diffusers}
HuggingFace.
\newblock Diffusers: State-of-the-art diffusion models.
\newblock \url{https://github.com/huggingface/diffusers}, 2023.

\bibitem[Jiang et~al.(2024)Jiang, Hu, Luo, He, Xu, Peng, Zhang, Wang, Wu, and Fu]{jiang2024fitdit}
Boyuan Jiang, Xiaobin Hu, Donghao Luo, Qingdong He, Chengming Xu, Jinlong Peng, Jiangning Zhang, Chengjie Wang, Yunsheng Wu, and Yanwei Fu.
\newblock Fitdit: Advancing the authentic garment details for high-fidelity virtual try-on.
\newblock \emph{arXiv preprint arXiv:2411.10499}, 2024.

\bibitem[Jin et~al.(2025)Jin, Peng, He, Hu, Chen, Wu, Zhu, Chi, Liu, Wang, et~al.]{jin2024dualanodiff}
Ying Jin, Jinlong Peng, Qingdong He, Teng Hu, Hao Chen, Jiafu Wu, Wenbing Zhu, Mingmin Chi, Jun Liu, Yabiao Wang, et~al.
\newblock Dualanodiff: Dual-interrelated diffusion model for few-shot anomaly image generation.
\newblock \emph{Proceedings of the IEEE/CVF conference on computer vision and pattern recognition}, 2025.

\bibitem[Ju et~al.(2024)Ju, Liu, Wang, Bian, Shan, and Xu]{brushnet}
Xuan Ju, Xian Liu, Xintao Wang, Yuxuan Bian, Ying Shan, and Qiang Xu.
\newblock Brushnet: A plug-and-play image inpainting model with decomposed dual-branch diffusion.
\newblock \emph{arXiv preprint arXiv:2403.06976}, 2024.

\bibitem[Kim et~al.(2024)Kim, Lee, Joung, Kim, and Baek]{instantfamily}
Chanran Kim, Jeongin Lee, Shichang Joung, Bongmo Kim, and Yeul-Min Baek.
\newblock Instantfamily: Masked attention for zero-shot multi-id image generation.
\newblock \emph{arXiv preprint arXiv:2404.19427}, 2024.

\bibitem[Kong et~al.(2025)Kong, Wu, Hu, Han, Peng, Xu, Luo, Zhang, Wang, and Fu]{kong2024anymaker}
Lingjie Kong, Kai Wu, Xiaobin Hu, Wenhui Han, Jinlong Peng, Chengming Xu, Donghao Luo, Jiangning Zhang, Chengjie Wang, and Yanwei Fu.
\newblock Anymaker: Zero-shot general object customization via decoupled dual-level id injection.
\newblock \emph{Proceedings of the IEEE/CVF conference on computer vision and pattern recognition}, 2025.

\bibitem[Kumari et~al.(2023)Kumari, Zhang, Zhang, Shechtman, and Zhu]{customDiffusion}
Nupur Kumari, Bingliang Zhang, Richard Zhang, Eli Shechtman, and Jun-Yan Zhu.
\newblock Multi-concept customization of text-to-image diffusion.
\newblock In \emph{Proceedings of the IEEE/CVF Conference on Computer Vision and Pattern Recognition}, pages 1931--1941, 2023.

\bibitem[Labs(2023)]{flux}
Black~Forest Labs.
\newblock Flux.
\newblock \url{https://github.com/black-forest-labs/flux}, 2023.

\bibitem[Li et~al.(2023{\natexlab{a}})Li, Li, and Hoi]{blipdiffusion}
Dongxu Li, Junnan Li, and Steven Hoi.
\newblock Blip-diffusion: Pre-trained subject representation for controllable text-to-image generation and editing.
\newblock \emph{Advances in Neural Information Processing Systems}, 36:\penalty0 30146--30166, 2023{\natexlab{a}}.

\bibitem[Li et~al.(2024)Li, Nie, Chen, Jiang, Wu, Lin, Liu, Peng, Wang, and Zheng]{li2024tuning}
Pengzhi Li, Qiang Nie, Ying Chen, Xi Jiang, Kai Wu, Yuhuan Lin, Yong Liu, Jinlong Peng, Chengjie Wang, and Feng Zheng.
\newblock Tuning-free image customization with image and text guidance.
\newblock In \emph{European Conference on Computer Vision}, pages 233--250. Springer, 2024.

\bibitem[Li et~al.(2023{\natexlab{b}})Li, Liu, Wu, Mu, Yang, Gao, Li, and Lee]{gligen}
Yuheng Li, Haotian Liu, Qingyang Wu, Fangzhou Mu, Jianwei Yang, Jianfeng Gao, Chunyuan Li, and Yong~Jae Lee.
\newblock Gligen: Open-set grounded text-to-image generation.
\newblock In \emph{Proceedings of the IEEE/CVF conference on computer vision and pattern recognition}, pages 22511--22521, 2023{\natexlab{b}}.

\bibitem[Lin et~al.(2025)Lin, Mo, Klingher, Mu, and Zhou]{ctrlx}
Kuan~Heng Lin, Sicheng Mo, Ben Klingher, Fangzhou Mu, and Bolei Zhou.
\newblock Ctrl-x: Controlling structure and appearance for text-to-image generation without guidance.
\newblock \emph{Advances in Neural Information Processing Systems}, 37:\penalty0 128911--128939, 2025.

\bibitem[Lipman et~al.(2022)Lipman, Chen, Ben-Hamu, Nickel, and Le]{fm}
Yaron Lipman, Ricky~TQ Chen, Heli Ben-Hamu, Maximilian Nickel, and Matt Le.
\newblock Flow matching for generative modeling.
\newblock \emph{arXiv preprint arXiv:2210.02747}, 2022.

\bibitem[Liu et~al.(2022)Liu, Gong, and Liu]{rf}
Xingchao Liu, Chengyue Gong, and Qiang Liu.
\newblock Flow straight and fast: Learning to generate and transfer data with rectified flow.
\newblock \emph{arXiv preprint arXiv:2209.03003}, 2022.

\bibitem[Mokady et~al.(2023)Mokady, Hertz, Aberman, Pritch, and Cohen-Or]{nulltextInversion}
Ron Mokady, Amir Hertz, Kfir Aberman, Yael Pritch, and Daniel Cohen-Or.
\newblock Null-text inversion for editing real images using guided diffusion models.
\newblock In \emph{Proceedings of the IEEE/CVF Conference on Computer Vision and Pattern Recognition}, pages 6038--6047, 2023.

\bibitem[Mou et~al.(2024)Mou, Wang, Xie, Wu, Zhang, Qi, and Shan]{T2Iadapter}
Chong Mou, Xintao Wang, Liangbin Xie, Yanze Wu, Jian Zhang, Zhongang Qi, and Ying Shan.
\newblock T2i-adapter: Learning adapters to dig out more controllable ability for text-to-image diffusion models.
\newblock In \emph{Proceedings of the AAAI Conference on Artificial Intelligence}, pages 4296--4304, 2024.

\bibitem[Peebles and Xie(2023)]{dit}
William Peebles and Saining Xie.
\newblock Scalable diffusion models with transformers.
\newblock In \emph{Proceedings of the IEEE/CVF International Conference on Computer Vision}, pages 4195--4205, 2023.

\bibitem[Peng et~al.(2024)Peng, Luo, Liu, and Zhang]{peng2024frih}
Jinlong Peng, Zekun Luo, Liang Liu, and Boshen Zhang.
\newblock Frih: fine-grained region-aware image harmonization.
\newblock In \emph{Proceedings of the AAAI Conference on Artificial Intelligence}, pages 4478--4486, 2024.

\bibitem[Podell et~al.(2023)Podell, English, Lacey, Blattmann, Dockhorn, M{\"u}ller, Penna, and Rombach]{sdxl}
Dustin Podell, Zion English, Kyle Lacey, Andreas Blattmann, Tim Dockhorn, Jonas M{\"u}ller, Joe Penna, and Robin Rombach.
\newblock Sdxl: Improving latent diffusion models for high-resolution image synthesis.
\newblock \emph{arXiv preprint arXiv:2307.01952}, 2023.

\bibitem[Qin et~al.(2023)Qin, Zhang, Yu, Feng, Yang, Zhou, Wang, Niebles, Xiong, Savarese, et~al.]{unicontrol}
Can Qin, Shu Zhang, Ning Yu, Yihao Feng, Xinyi Yang, Yingbo Zhou, Huan Wang, Juan~Carlos Niebles, Caiming Xiong, Silvio Savarese, et~al.
\newblock Unicontrol: A unified diffusion model for controllable visual generation in the wild.
\newblock \emph{arXiv preprint arXiv:2305.11147}, 2023.

\bibitem[Radford et~al.(2021)Radford, Kim, Hallacy, Ramesh, Goh, Agarwal, Sastry, Askell, Mishkin, Clark, et~al.]{clip}
Alec Radford, Jong~Wook Kim, Chris Hallacy, Aditya Ramesh, Gabriel Goh, Sandhini Agarwal, Girish Sastry, Amanda Askell, Pamela Mishkin, Jack Clark, et~al.
\newblock Learning transferable visual models from natural language supervision.
\newblock In \emph{International conference on machine learning}, pages 8748--8763. PMLR, 2021.

\bibitem[Rombach et~al.(2022)Rombach, Blattmann, Lorenz, Esser, and Ommer]{ldm}
Robin Rombach, Andreas Blattmann, Dominik Lorenz, Patrick Esser, and Bj{\"o}rn Ommer.
\newblock High-resolution image synthesis with latent diffusion models.
\newblock In \emph{Proceedings of the IEEE/CVF conference on computer vision and pattern recognition}, pages 10684--10695, 2022.

\bibitem[Ronneberger et~al.(2015)Ronneberger, Fischer, and Brox]{unet}
Olaf Ronneberger, Philipp Fischer, and Thomas Brox.
\newblock U-net: Convolutional networks for biomedical image segmentation.
\newblock In \emph{Medical image computing and computer-assisted intervention--MICCAI 2015: 18th international conference, Munich, Germany, October 5-9, 2015, proceedings, part III 18}, pages 234--241. Springer, 2015.

\bibitem[Rout et~al.(2024)Rout, Chen, Ruiz, Caramanis, Shakkottai, and Chu]{RFinversion}
Litu Rout, Yujia Chen, Nataniel Ruiz, Constantine Caramanis, Sanjay Shakkottai, and Wen-Sheng Chu.
\newblock Semantic image inversion and editing using rectified stochastic differential equations.
\newblock \emph{arXiv preprint arXiv:2410.10792}, 2024.

\bibitem[Ruiz et~al.(2023)Ruiz, Li, Jampani, Pritch, Rubinstein, and Aberman]{dreambooth}
Nataniel Ruiz, Yuanzhen Li, Varun Jampani, Yael Pritch, Michael Rubinstein, and Kfir Aberman.
\newblock Dreambooth: Fine tuning text-to-image diffusion models for subject-driven generation.
\newblock In \emph{Proceedings of the IEEE/CVF conference on computer vision and pattern recognition}, pages 22500--22510, 2023.

\bibitem[Sohn et~al.(2023)Sohn, Ruiz, Lee, Chin, Blok, Chang, Barber, Jiang, Entis, Li, et~al.]{styledrop}
Kihyuk Sohn, Nataniel Ruiz, Kimin Lee, Daniel~Castro Chin, Irina Blok, Huiwen Chang, Jarred Barber, Lu Jiang, Glenn Entis, Yuanzhen Li, et~al.
\newblock Styledrop: Text-to-image generation in any style.
\newblock \emph{arXiv preprint arXiv:2306.00983}, 2023.

\bibitem[Song et~al.(2020)Song, Meng, and Ermon]{ddim}
Jiaming Song, Chenlin Meng, and Stefano Ermon.
\newblock Denoising diffusion implicit models.
\newblock \emph{arXiv preprint arXiv:2010.02502}, 2020.

\bibitem[Song et~al.(2022)Song, Zhang, Lin, Cohen, Price, Zhang, Kim, and Aliaga]{objectStitch}
Yizhi Song, Zhifei Zhang, Zhe Lin, Scott Cohen, Brian Price, Jianming Zhang, Soo~Ye Kim, and Daniel Aliaga.
\newblock Objectstitch: Generative object compositing.
\newblock \emph{arXiv preprint arXiv:2212.00932}, 2022.

\bibitem[Su et~al.(2024)Su, Ahmed, Lu, Pan, Bo, and Liu]{rotaryPE}
Jianlin Su, Murtadha Ahmed, Yu Lu, Shengfeng Pan, Wen Bo, and Yunfeng Liu.
\newblock Roformer: Enhanced transformer with rotary position embedding.
\newblock \emph{Neurocomputing}, 568:\penalty0 127063, 2024.

\bibitem[Tan et~al.(2024)Tan, Liu, Yang, Xue, and Wang]{ominicontrol}
Zhenxiong Tan, Songhua Liu, Xingyi Yang, Qiaochu Xue, and Xinchao Wang.
\newblock Ominicontrol: Minimal and universal control for diffusion transformer.
\newblock \emph{arXiv preprint arXiv:2411.15098}, 3, 2024.

\bibitem[Wang et~al.(2025)Wang, He, Peng, Yang, Chi, and Wang]{mambayoloworld}
Haoxuan Wang, Qingdong He, Jinlong Peng, Hao Yang, Mingmin Chi, and Yabiao Wang.
\newblock Mamba-yolo-world: Marrying yolo-world with mamba for open-vocabulary detection.
\newblock \emph{IEEE International Conference on Acoustics, Speech, and Signal Processing}, 2025.

\bibitem[Wang et~al.(2024{\natexlab{a}})Wang, Bai, Wang, Qin, Chen, Li, Tang, and Hu]{instantid}
Qixun Wang, Xu Bai, Haofan Wang, Zekui Qin, Anthony Chen, Huaxia Li, Xu Tang, and Yao Hu.
\newblock Instantid: Zero-shot identity-preserving generation in seconds.
\newblock \emph{arXiv preprint arXiv:2401.07519}, 2024{\natexlab{a}}.

\bibitem[Wang et~al.(2023)Wang, Saharia, Montgomery, Pont-Tuset, Noy, Pellegrini, Onoe, Laszlo, Fleet, Soricut, et~al.]{imageneditor}
Su Wang, Chitwan Saharia, Ceslee Montgomery, Jordi Pont-Tuset, Shai Noy, Stefano Pellegrini, Yasumasa Onoe, Sarah Laszlo, David~J Fleet, Radu Soricut, et~al.
\newblock Imagen editor and editbench: Advancing and evaluating text-guided image inpainting.
\newblock In \emph{Proceedings of the IEEE/CVF conference on computer vision and pattern recognition}, pages 18359--18369, 2023.

\bibitem[Wang et~al.(2024{\natexlab{b}})Wang, Darrell, Rambhatla, Girdhar, and Misra]{instancediffusion}
Xudong Wang, Trevor Darrell, Sai~Saketh Rambhatla, Rohit Girdhar, and Ishan Misra.
\newblock Instancediffusion: Instance-level control for image generation, 2024{\natexlab{b}}.

\bibitem[Wang et~al.(2004)Wang, Bovik, Sheikh, and Simoncelli]{ssim}
Zhou Wang, Alan~C Bovik, Hamid~R Sheikh, and Eero~P Simoncelli.
\newblock Image quality assessment: from error visibility to structural similarity.
\newblock \emph{IEEE transactions on image processing}, 13\penalty0 (4):\penalty0 600--612, 2004.

\bibitem[Winter et~al.(2024)Winter, Shul, Cohen, Berman, Pritch, Rav-Acha, and Hoshen]{objectmate}
Daniel Winter, Asaf Shul, Matan Cohen, Dana Berman, Yael Pritch, Alex Rav-Acha, and Yedid Hoshen.
\newblock Objectmate: A recurrence prior for object insertion and subject-driven generation.
\newblock \emph{arXiv preprint arXiv:2412.08645}, 2024.

\bibitem[Xing et~al.(2024)Xing, Wang, Sun, Wang, Bai, Ai, Huang, and Li]{csgo}
Peng Xing, Haofan Wang, Yanpeng Sun, Qixun Wang, Xu Bai, Hao Ai, Renyuan Huang, and Zechao Li.
\newblock Csgo: Content-style composition in text-to-image generation.
\newblock \emph{arXiv preprint arXiv:2408.16766}, 2024.

\bibitem[XLabs-AI(2024{\natexlab{a}})]{XLabs-AI_Flux-ControlNet-Canny}
XLabs-AI.
\newblock Flux-controlnet-canny-diffusers.
\newblock \url{https://huggingface.co/XLabs-AI/flux-controlnet-canny-diffusers}, 2024{\natexlab{a}}.

\bibitem[XLabs-AI(2024{\natexlab{b}})]{XLabs-AI_Flux-ControlNet-Depth}
XLabs-AI.
\newblock Flux-controlnet-depth-diffusers.
\newblock \url{https://huggingface.co/XLabs-AI/flux-controlnet-depth-diffusers}, 2024{\natexlab{b}}.

\bibitem[XLabs-AI(2024{\natexlab{c}})]{XLabs-AI_Flux-ip-adapter}
XLabs-AI.
\newblock Flux-ip-adapter.
\newblock \url{https://huggingface.co/XLabs-AI/flux-ip-adapter}, 2024{\natexlab{c}}.

\bibitem[Yang et~al.(2023)Yang, Gu, Zhang, Zhang, Chen, Sun, Chen, and Wen]{pbe}
Binxin Yang, Shuyang Gu, Bo Zhang, Ting Zhang, Xuejin Chen, Xiaoyan Sun, Dong Chen, and Fang Wen.
\newblock Paint by example: Exemplar-based image editing with diffusion models.
\newblock In \emph{Proceedings of the IEEE/CVF Conference on Computer Vision and Pattern Recognition}, pages 18381--18391, 2023.

\bibitem[Yang et~al.(2024)Yang, Kang, Huang, Xu, Feng, and Zhao]{depthanything}
Lihe Yang, Bingyi Kang, Zilong Huang, Xiaogang Xu, Jiashi Feng, and Hengshuang Zhao.
\newblock Depth anything: Unleashing the power of large-scale unlabeled data.
\newblock In \emph{CVPR}, 2024.

\bibitem[Ye et~al.(2023)Ye, Zhang, Liu, Han, and Yang]{IPadapter}
Hu Ye, Jun Zhang, Sibo Liu, Xiao Han, and Wei Yang.
\newblock Ip-adapter: Text compatible image prompt adapter for text-to-image diffusion models.
\newblock \emph{arXiv preprint arXiv:2308.06721}, 2023.

\bibitem[Zhang et~al.(2023)Zhang, Rao, and Agrawala]{controlnet}
Lvmin Zhang, Anyi Rao, and Maneesh Agrawala.
\newblock Adding conditional control to text-to-image diffusion models.
\newblock In \emph{Proceedings of the IEEE/CVF International Conference on Computer Vision}, pages 3836--3847, 2023.

\bibitem[Zhao et~al.(2024)Zhao, Chen, Chen, Bao, Hao, Yuan, and Wong]{unicontrolnet}
Shihao Zhao, Dongdong Chen, Yen-Chun Chen, Jianmin Bao, Shaozhe Hao, Lu Yuan, and Kwan-Yee~K Wong.
\newblock Uni-controlnet: All-in-one control to text-to-image diffusion models.
\newblock \emph{Advances in Neural Information Processing Systems}, 36, 2024.

\bibitem[Zhuang et~al.(2025)Zhuang, Zeng, Liu, Yuan, and Chen]{powerpaint}
Junhao Zhuang, Yanhong Zeng, Wenran Liu, Chun Yuan, and Kai Chen.
\newblock A task is worth one word: Learning with task prompts for high-quality versatile image inpainting.
\newblock In \emph{European Conference on Computer Vision}, pages 195--211. Springer, 2025.

\end{thebibliography}
